\documentclass[letterpaper, 10 pt, conference]{ieeeconf}  
\let\labelindent\relax
\usepackage[loadonly]{enumitem}

\IEEEoverridecommandlockouts                              

\usepackage{graphics} 
\usepackage{epsfig} 
\usepackage{mathptmx} 
\usepackage{times} 
\usepackage{amsmath} 
\usepackage{amssymb}  
\usepackage[T1]{fontenc}
\usepackage[utf8]{inputenc}
\usepackage{lmodern}  
\usepackage[table]{xcolor} 

\usepackage{graphicx}             
\usepackage{caption}              
\usepackage{subcaption}           

\usepackage{booktabs}             
\usepackage{multirow}             
\usepackage{array}                
\usepackage{tabularx}             
\usepackage{float}                

\usepackage{algorithm}            
\usepackage{algorithmic}          

\usepackage{enumitem}             

\usepackage{xcolor}               
\usepackage{float}

\usepackage{url}
\usepackage{cite}
\usepackage[hidelinks]{hyperref}
\usepackage{cleveref}
\phantomsection  

\title{\LARGE \bf
ADAM-Dehaze: Adaptive
Density-Aware Multi-Stage Dehazing
for Improved Object Detection in
Foggy Conditions
}

\author{Fatmah AlHindaassi$^{1^*}$, Mohammed Talha Alam$^{1^*}$ and Fakhri Karray$^{1}$
\thanks{*Equal Contribution}
\thanks{$^{1}$Mohamed bin Zayed University of Artificial Intelligence, UAE
        {\tt\small mohammed.alam@mbzuai.ac.ae;}
        Appendix available \href{https://www.dropbox.com/scl/fi/mk44qb2ql3442gf71clw3/Appendix.pdf?rlkey=b802puu0etvj9y1yhii1in5tm&dl=0} {\texttt{here}.}}%
}

\begin{document}

\maketitle
\thispagestyle{empty}
\pagestyle{empty}

\begin{abstract}

Adverse weather conditions, particularly fog, pose a significant challenge to autonomous vehicles, surveillance systems, and other safety-critical applications by severely degrading visual information. We introduce \textit{ADAM‑Dehaze}, an adaptive, density‑aware dehazing framework that jointly optimizes image restoration and object detection under varying fog intensities. First, a lightweight Haze Density Estimation Network (HDEN) classifies each input as light, medium, or heavy fog. Based on this score, the system dynamically routes the image through one of three CORUN branches—Light, Medium, or Complex—each tailored to its haze regime. A novel adaptive loss then balances physical‐model coherence and perceptual fidelity, ensuring both accurate defogging and preservation of fine details. On Cityscapes and the real‑world RTTS benchmark, ADAM‑Dehaze boosts PSNR by up to 2.1 dB, reduces FADE by 30\%, and improves object detection mAP by up to 13 points, all while cutting inference time by 20\%. These results demonstrate the necessity of intensity‑specific processing and seamless integration with downstream vision tasks for robust performance in foggy weather conditions.


\end{abstract}

\section{INTRODUCTION}

Computer vision powers critical systems—from autonomous vehicles and traffic monitoring to environmental sensing and forensic inspection—yet its performance still collapses in fog \cite{narasimhan2002vision}. By scattering and absorbing light, fog reduces contrast, obscures details, and distorts colors \cite{he2010single, li2018benchmarking}, causing standard detectors (trained on clear scenes) to miss objects or generate false alarms \cite{michaelis2019benchmarking, redmon2018yolov3, ren2015faster}. In safety‑critical domains such as self‑driving cars and intelligent transportation systems, even small reliability drops can have catastrophic consequences \cite{sakaridis2018model, hahner2019semantic}.

Classic dehazing methods like the Dark Channel Prior (DCP) exploit physical models to partially restore clarity \cite{he2010single, lee2016review}, but they often leave residual haze or introduce halos, especially in dense fog \cite{zhang2018saliency, li2017haze}. Deep‑learning approaches yield better perceptual quality but typically apply a uniform, fixed‐capacity network regardless of actual fog density \cite{dong2020multi, ren2018gated, li2017aodnet}. This “one‑size‑fits‑all” strategy wastes computation on lightly fogged images and underperforms on heavily obscured ones, leading to inconsistent downstream detection results \cite{li2018end, zhang2020unified}.

Joint dehazing–detection pipelines have shown that co‑optimizing restoration and recognition can recover both image quality and task accuracy \cite{li2018end, zhang2020unified}, but existing works generally treat dehazing as a monolithic preprocessing step. They do not exploit the insight that optimal dehazing parameters—and indeed optimal network capacity—vary dramatically with fog intensity \cite{ye2022perceiving, wu2023ridcp, chen2021psd}. Nor do they adapt their training losses to enforce a fog‑aware balance between physical consistency, perceptual realism, and task‑driven feature preservation \cite{yang2022self, mou2022deep}.

\begin{figure}
    \centering
    \includegraphics[width=1.0\linewidth]{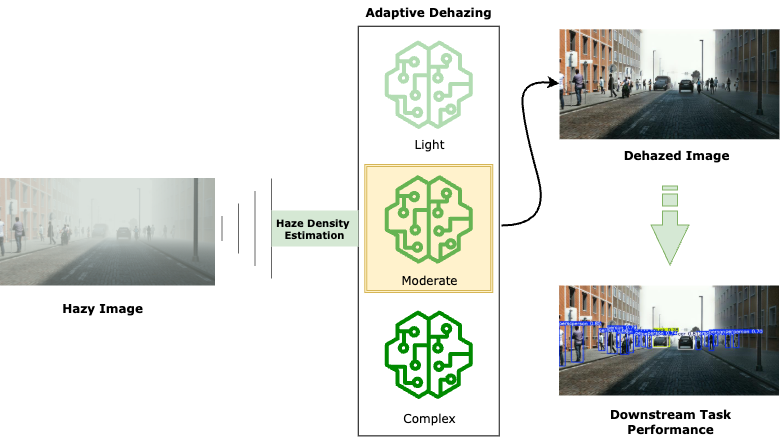}
    \caption{Conceptual illustration of ADAM-Dehaze. Given a foggy input image, our framework adaptively estimates the haze density and selects a specialized dehazing branch—light, moderate, or complex—accordingly.}
    \label{fig:concept}
\end{figure}


To address these limitations, we present \textit{ADAM-Dehaze}, an \textbf{A}daptive, \textbf{D}ensity-\textbf{A}ware, \textbf{M}ulti-stage dehazing framework that dynamically tailors both network complexity and loss design to the scene’s fog intensity. At its core is a lightweight Haze Density Estimation Network (HDEN) that classifies input images into three regimes—light, medium, or heavy fog. Based on this estimate, ADAM‑Dehaze routes each image through one of three specialized unfolding networks (CORUN‑Light, CORUN‑Medium, or CORUN‑Complex), ensuring minimal resource use for easy cases and maximum power for the hardest ones. A density‑modulated loss then balances physical‐model coherence, perceptual fidelity, and haze‐consistency, yielding clear outputs without over‑ or undercorrection \cite{qin2020ffa, mou2022deep}.

On both synthetic Cityscapes \cite{cordts2016cityscapes} and the real‑world RTTS benchmark  \cite{li2018benchmarking}, ADAM‑Dehaze consistently improves perceptual quality and boosts object‐detection mAP—particularly under heavy fog—while reducing average inference time. Our results highlight the value of coupling fog density estimation with adaptive dehazing and downstream task integration. Based on this idea, we make the following contributions in this work:
\begin{itemize}
    \item \textit{FogIntensity-25K dataset}. We introduce and will release \textbf{FogIntensity-25K}, a synthetic dataset stratified into light, medium, and heavy fog categories based on the atmospheric scattering model. It provides paired hazy/clear images with depth-based fog simulation, facilitating future research on intensity-aware restoration and fog-adaptive vision tasks.
    \item \textit{Adaptive dehazing network}s. Three CORUN variants (2/4/6 stages + attention) optimized for light, medium, and heavy fog regimes.
    \item \textit{Modulated loss function}. A fog‐intensity–weighted combination of physical‐model coherence, perceptual, and density terms.
    \item \textit{Improved detection under fog}. Demonstration of up to +13 mAP gain on real‐world foggy images with 20\% faster inference.
\end{itemize}

\begin{figure*}[htp]
    \centering
    \includegraphics[width=0.90\textwidth]{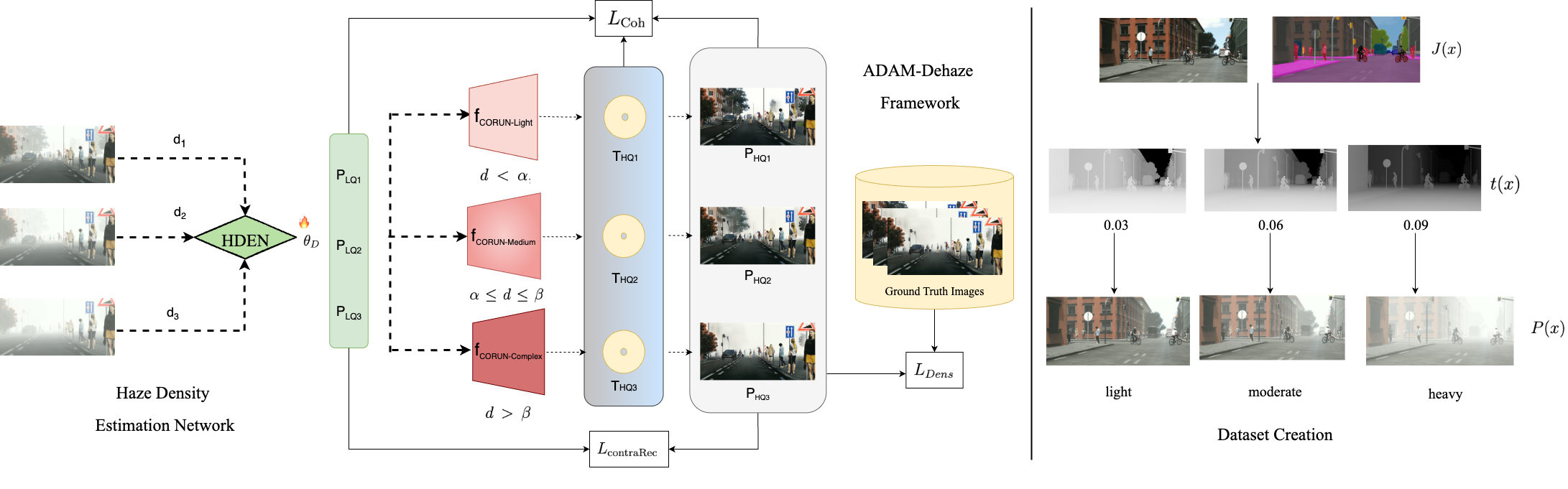}
    \caption{Architecture of the proposed ADAM-Dehaze framework. The process begins with generating foggy images at various haze intensities from clear-weather images using corresponding depth maps. Three haze intensity levels ($\beta$ = 0.03 light, 0.06 medium, 0.09 complex) are selected and classified by HDEN, each subsequently processed by a dedicated dehazing network. The final dehazed outputs are then passed to YOLOv8n for robust object detection under varying fog conditions.}
    \label{fig:Flowchart}
\end{figure*}
\section{RELATED WORK}
\textbf{Single‑Image Dehazing.} Early methods for image dehazing relied on physical priors imposed on the atmospheric scattering model. He et al. introduced the Dark Channel Prior (DCP) \cite{he2010single}, exploiting the observation that clear outdoor images contain at least one color channel with very low intensity in haze‑free regions. While DCP often improves contrast in moderate haze, it can overestimate transmission in bright areas (e.g., sky, white objects), producing halos and color shifts \cite{tanya2021enhanced,riaz2016single}. With the advent of deep learning, end‑to‑end networks learned to directly map hazy to clear images \cite{lihe2024phdnet, satrasupalli2020end, cai2016dehazenet}. Dong et al. proposed MSBDN \cite{dong2020multi}, a multi‑scale boosted dehazing network that fuses dense feature representations across scales, significantly improving PSNR and SSIM on synthetic benchmarks. GridDehazeNet introduced attention‑guided multi‑scale feature fusion \cite{liu2019griddehazenet}, and FFA‑Net combined feature fusion with spatial and channel attention to better recover fine textures \cite{qin2020ffa}. Although these methods excel in quantitative metrics, they assume a fixed network capacity regardless of haze severity, leading to under‑ or over‑processing in light or heavy fog.

\textbf{Deep Unfolding and Intensity‑Aware Dehazing.} To better integrate physical models, deep unfolding networks (DUNs) embed iterative optimization steps into neural architectures \cite{qin2023accelerated, yang2022self, adam2024optimizing}. More recently, CORUN \cite{dong2020multi} introduced cooperative unfolding modules to jointly optimize scene and transmission estimates, achieving state‑of‑the‑art restoration. Intensity‑aware methods explicitly tailor processing to fog density: Yang et al.’s PSD \cite{chen2021psd} adjusts restoration strength via local transmission, RIDCP \cite{wu2023ridcp} employs a learned codebook of high‑quality priors and a Controllable HQPs matching step for domain adaptation. These approaches highlight the benefit of density‑guided dehazing but do not jointly optimize downstream tasks. Extending these model‑based and density‑adaptive paradigms, diffusion‑driven augmentation frameworks such as FLARE \cite{alam2024flare} leverage a denoising diffusion process to generate class‑balanced, high‑resolution variants of hazy inputs, realigning feature distributions and substantially boosting downstream performance.

\textbf{Object Detection under Fog.} Standard detectors—YOLOv3 \cite{redmon2018yolov3}, Faster R‑CNN  \cite{ren2015faster}, SSD \cite{liu2016ssd}—suffer steep performance drops when applied to foggy inputs. AOD‑Net \cite{li2017aodnet} couples a lightweight dehazing module with YOLOv5s, partially restoring detection accuracy in light haze, but its uniform design underperforms in heavy fog. Domain‑adapted Faster R‑CNN variants (e.g., DA‑FRCNN) and joint segmentation–detection frameworks \cite{zhang2020unified} mitigate some domain shift but lack explicit dehazing control. Similarly, unified frameworks that include detection losses guide the network toward downstream performance but still treat dehazing as a monolithic process.

\section{Methodology}

Our proposed \textbf{ADAM-Dehaze} framework (\textbf{A}daptive \textbf{D}ensity-\textbf{A}ware \textbf{M}ulti-stage Dehazing) addresses the limitations of fixed-capacity dehazing networks by tailoring restoration complexity to the severity of fog. The architecture dynamically modulates both the processing path and the training objective based on a predicted haze density score. This section outlines our dataset construction strategy and the core components of the ADAM-Dehaze pipeline.

\subsection{Dataset Generation}

Robust dehazing and detection require both perceptual fidelity and real-world generalization. We therefore construct a hybrid dataset combining synthetic clear–hazy pairs and uncontrolled real-world foggy scenes:

\subsubsection{Synthetic FogIntensity-25K dataset}

We synthesize 25,000 paired hazy images using the Atmospheric Scattering Model (ASM), applied to Cityscapes and Synscapes datasets:
\begin{equation}
P(x) = J(x) \cdot t(x) + A(1 - t(x)),
\label{eq:asm}
\end{equation}
where \( P(x) \) is the observed hazy image, \( J(x) \) is the clear counterpart, \( A \) is the global atmospheric light, and \( t(x) \) is the transmission map:
\begin{equation}
t(x) = \exp(-\beta d(x)),
\label{eq:trans}
\end{equation}
with \( d(x) \) representing scene depth and \( \beta \in \{0.03, 0.06, 0.09\} \) controlling light, medium, and heavy fog densities. Depth-aware attenuation ensures distant objects appear more obscured, mimicking real-world fog effects. These paired samples enable full-reference evaluation using PSNR, SSIM, and LPIPS.

\subsubsection{Real-World Foggy Images}

We use the RTTS dataset (approx. 4,000 images) for evaluation in uncontrolled foggy conditions. Since ground-truth clear images are unavailable, we employ non-reference image quality metrics such as FADE \cite{choi2015referenceless}, BRISQUE \cite{mittal2012no}, and NIMA \cite{talebi2018nima}.

This hybrid data construction supports comprehensive training and evaluation across simulated and real-world domains.

\begin{table}[htp]
\centering
\caption{FogIntensity-25K dataset composition.}
\small
\begin{tabular}{lccp{3.5cm}}
\toprule
\textbf{Fog Level} & \textbf{$\beta$} & \textbf{\#Images} & \textbf{Description} \\
\midrule
\rowcolor[HTML]{F2F2F2}
Light  & 0.03 & 8,333 & Mild degradation, visibility preserved \\
\rowcolor[HTML]{F6FBFF}
Medium & 0.06 & 8,333 & Moderate degradation \\
\rowcolor[HTML]{EFF6FC}
Heavy  & 0.09 & 8,334 & Severe visibility loss \\
\bottomrule
\end{tabular}
\end{table}

\begin{figure}[H]  
    \centering
    \setlength{\tabcolsep}{2pt} 
    \begin{tabular}{cccccc}
        \includegraphics[width=0.23\linewidth]{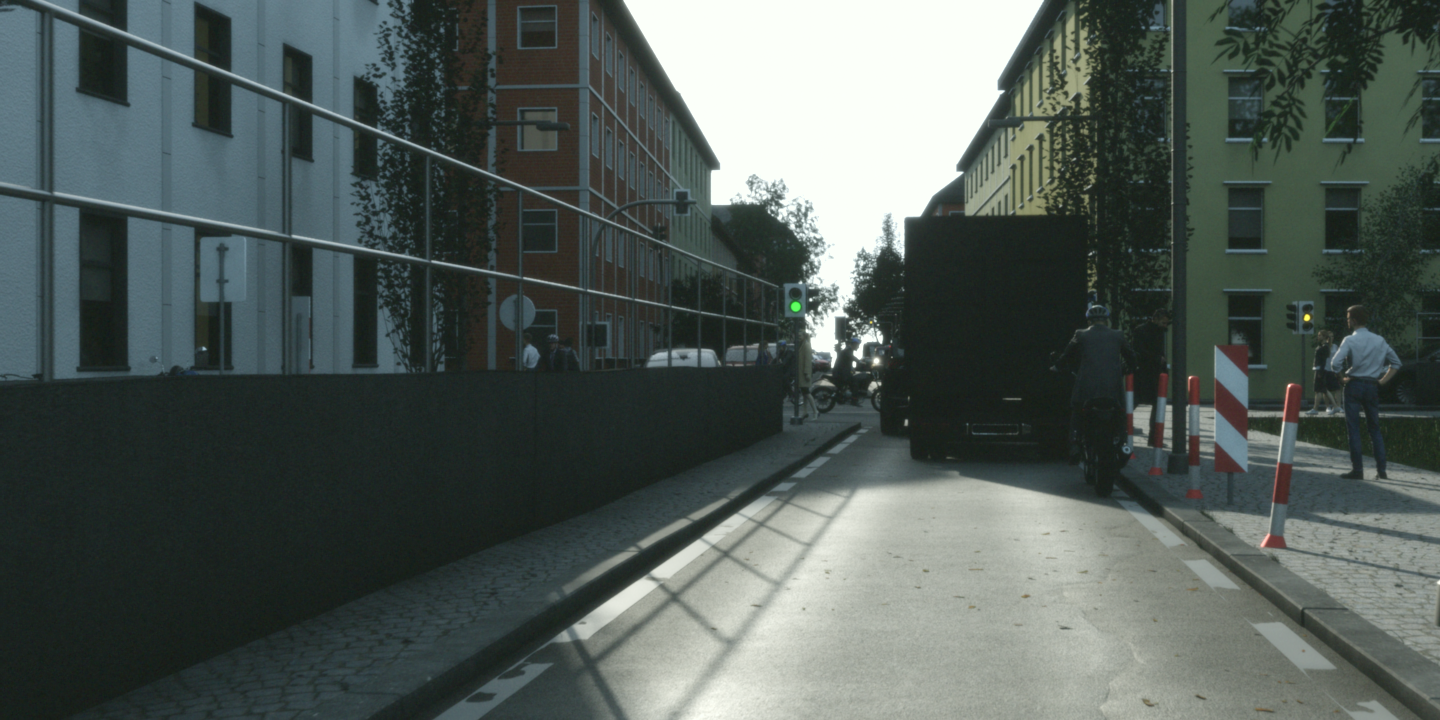} &
        \includegraphics[width=0.23\linewidth]{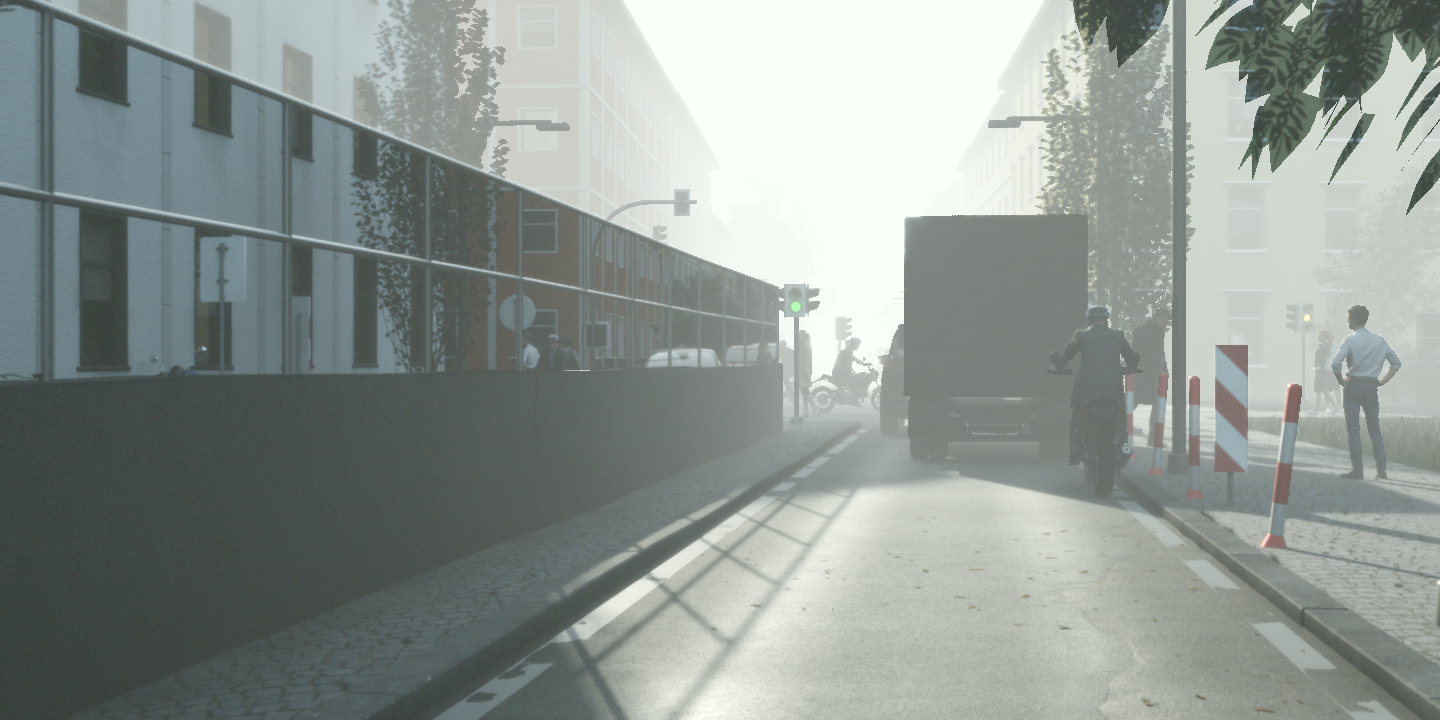} &
        \includegraphics[width=0.23\linewidth]{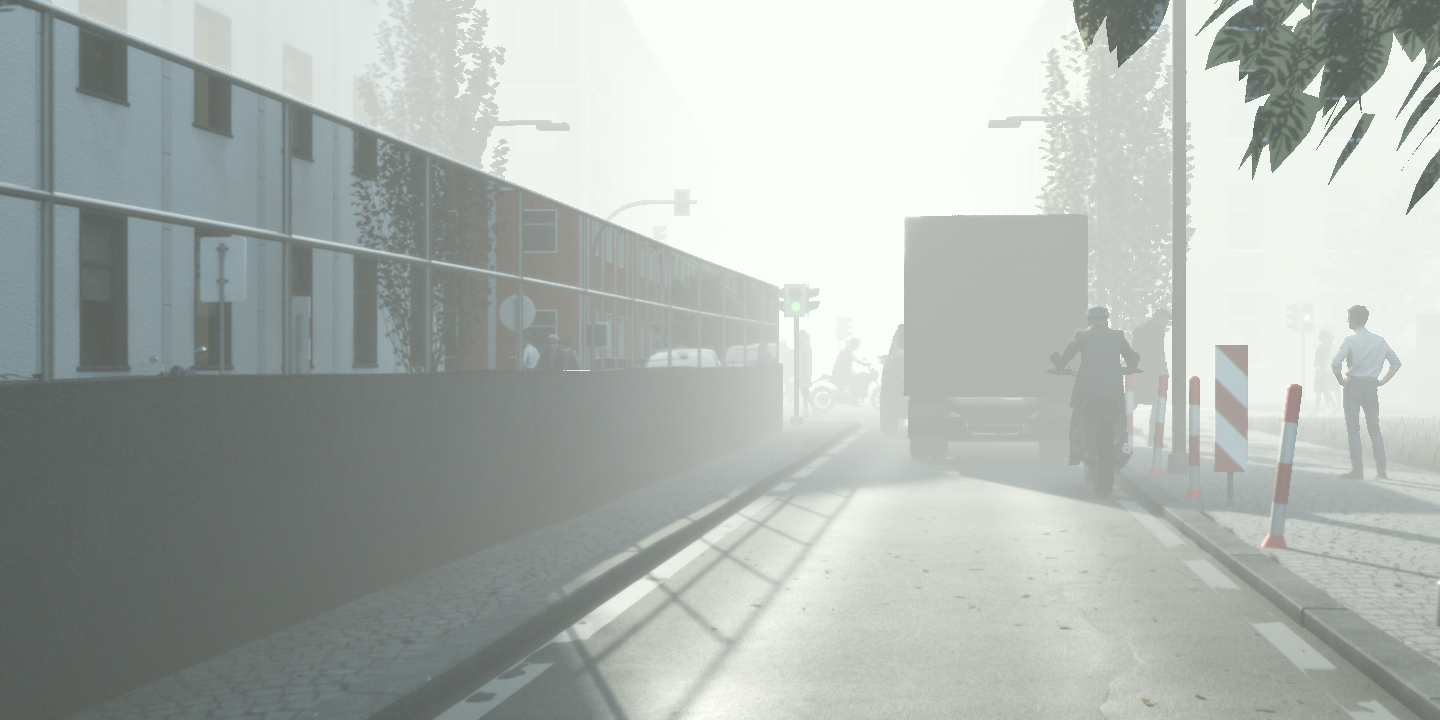} &
        \includegraphics[width=0.23\linewidth]{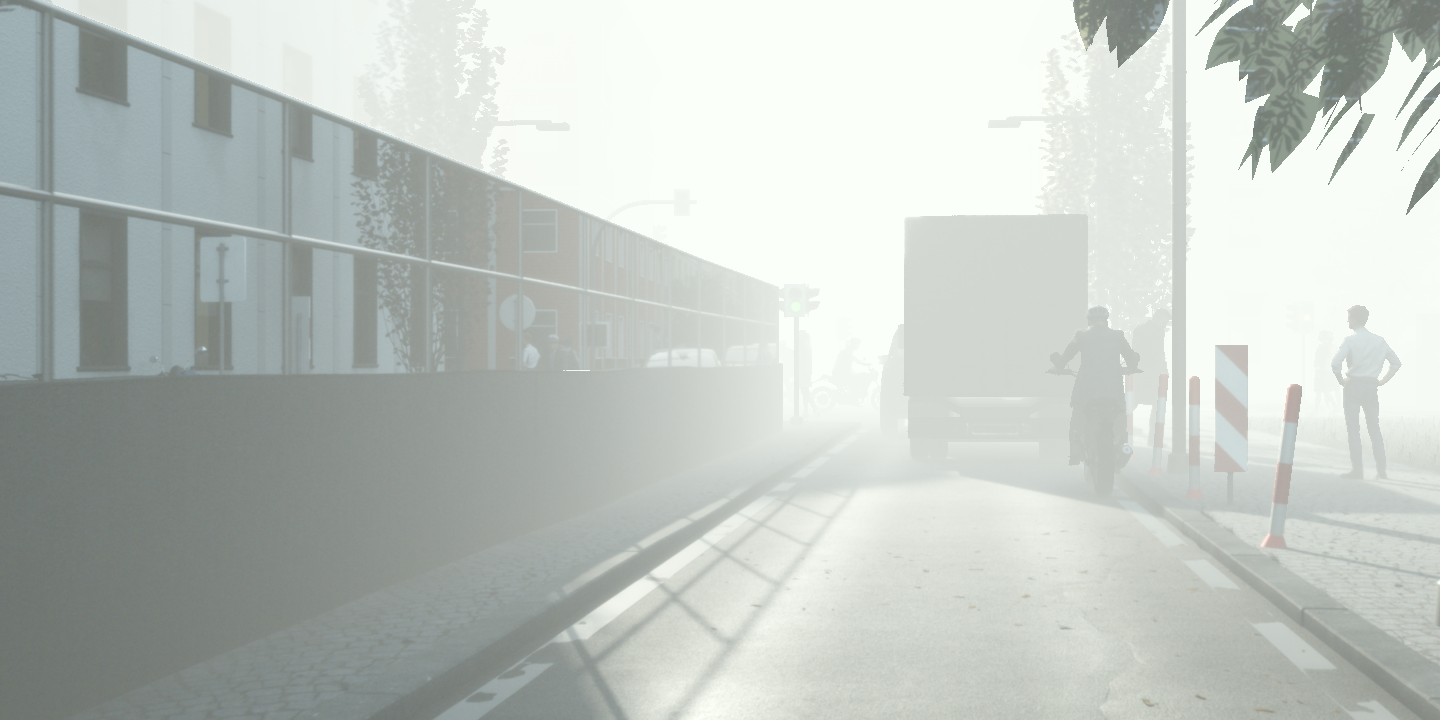} \\
        
        \includegraphics[width=0.23\linewidth]{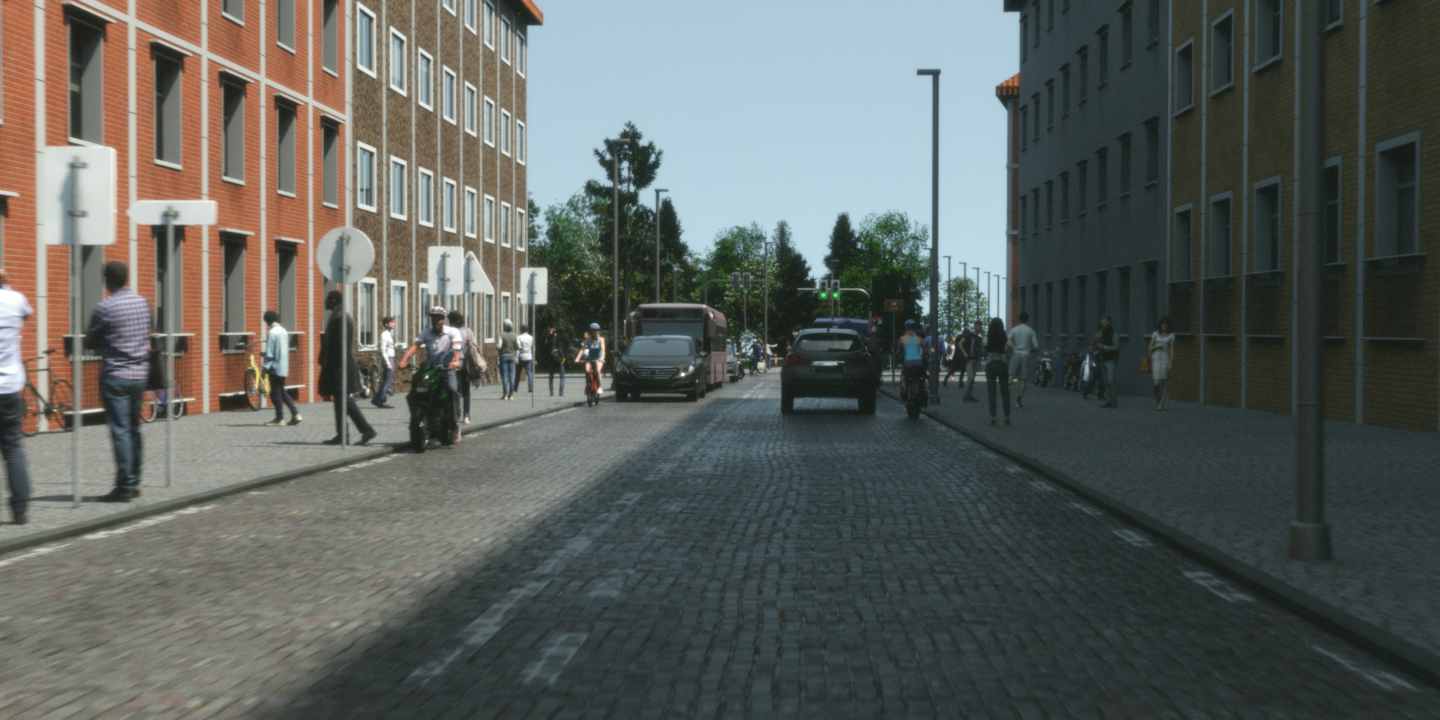} &
        \includegraphics[width=0.23\linewidth]{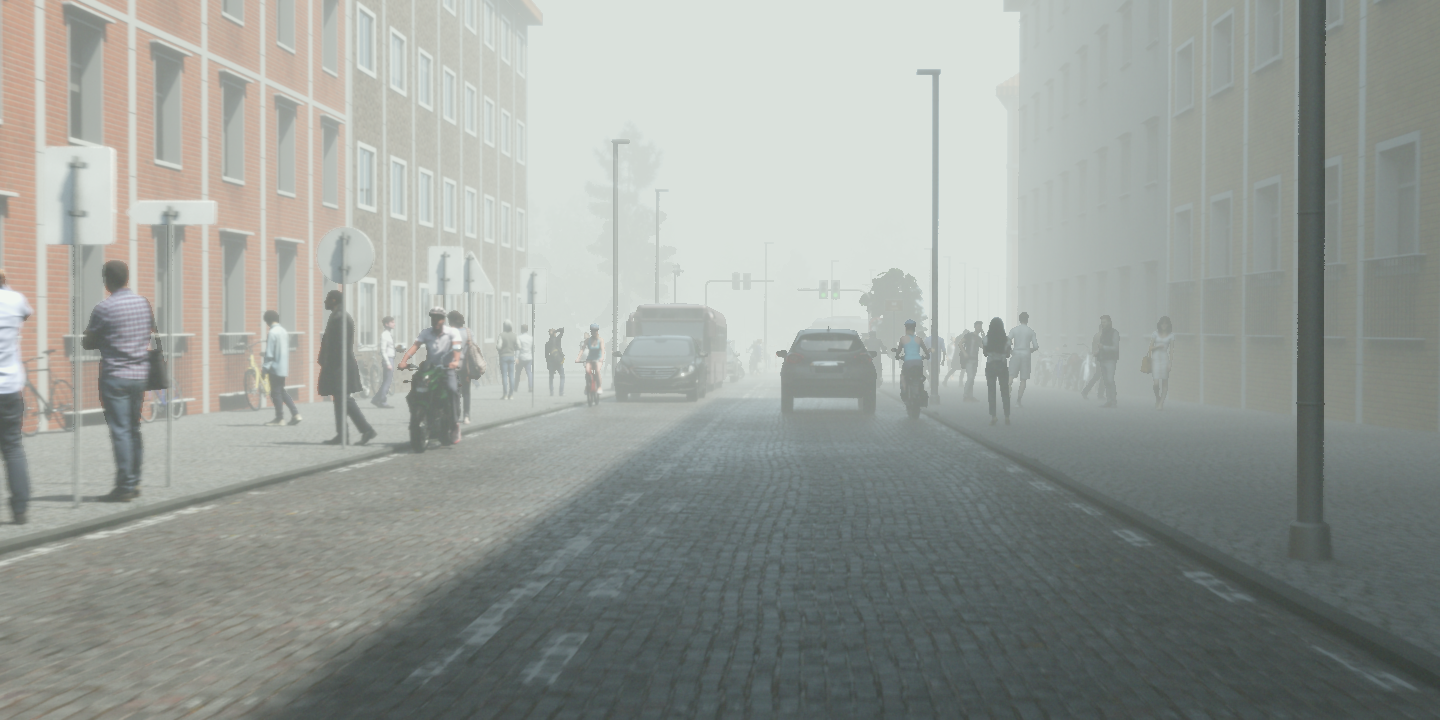} &
        \includegraphics[width=0.23\linewidth]{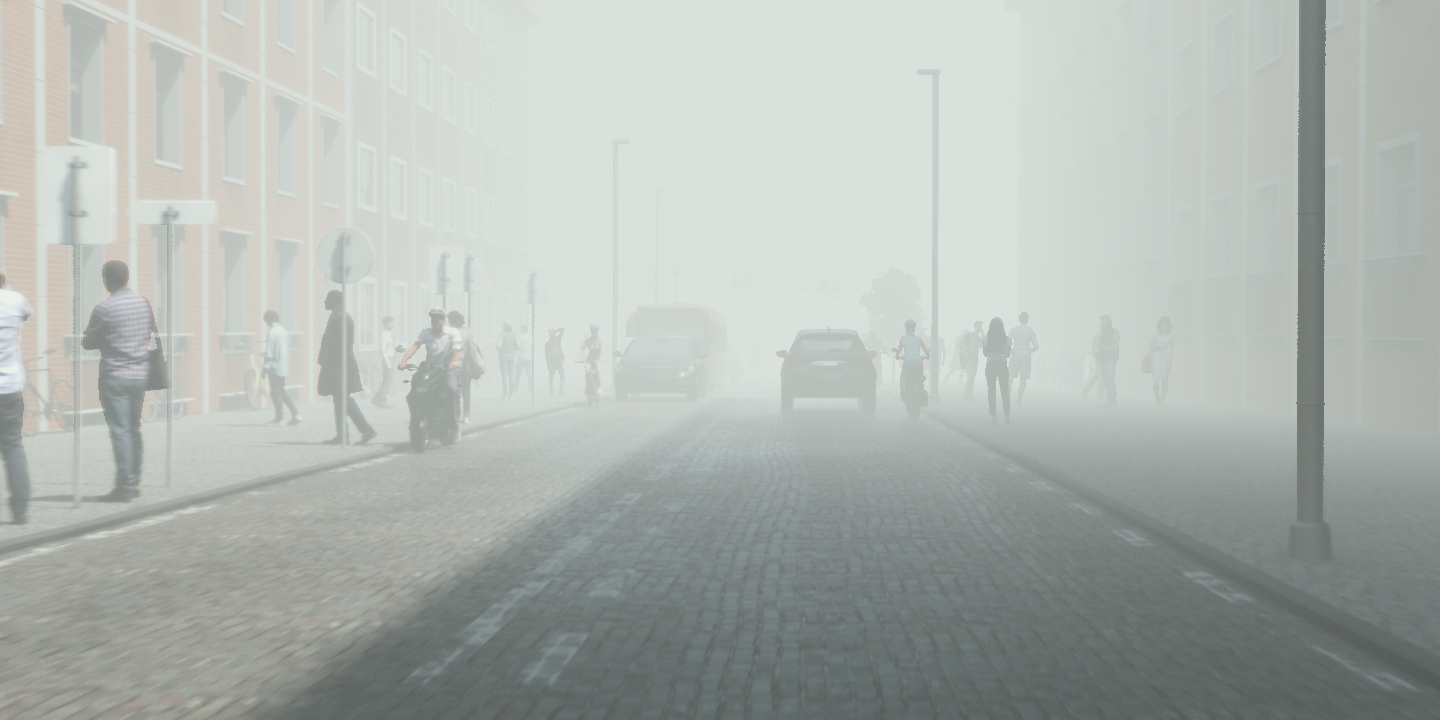} &
        \includegraphics[width=0.23\linewidth]{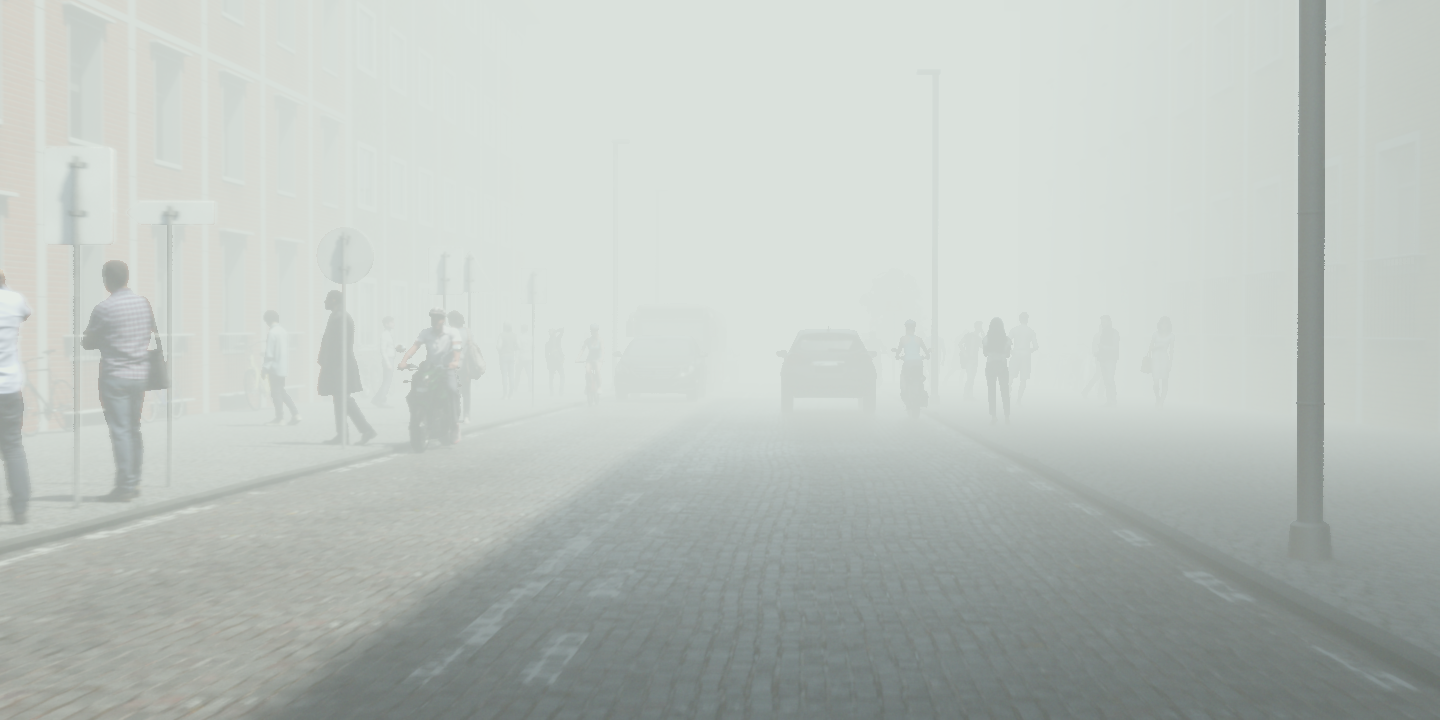}\\
        
        \includegraphics[width=0.23\linewidth]{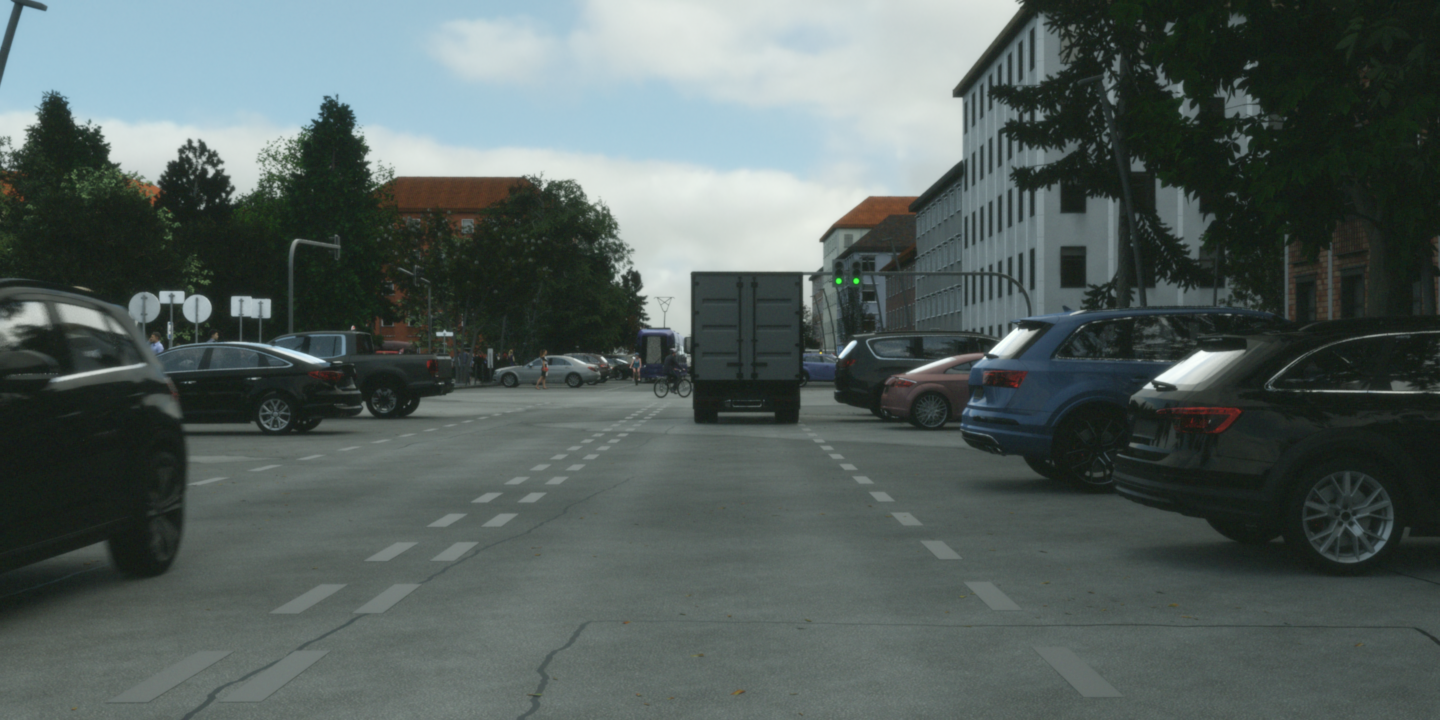} &
        \includegraphics[width=0.23\linewidth]{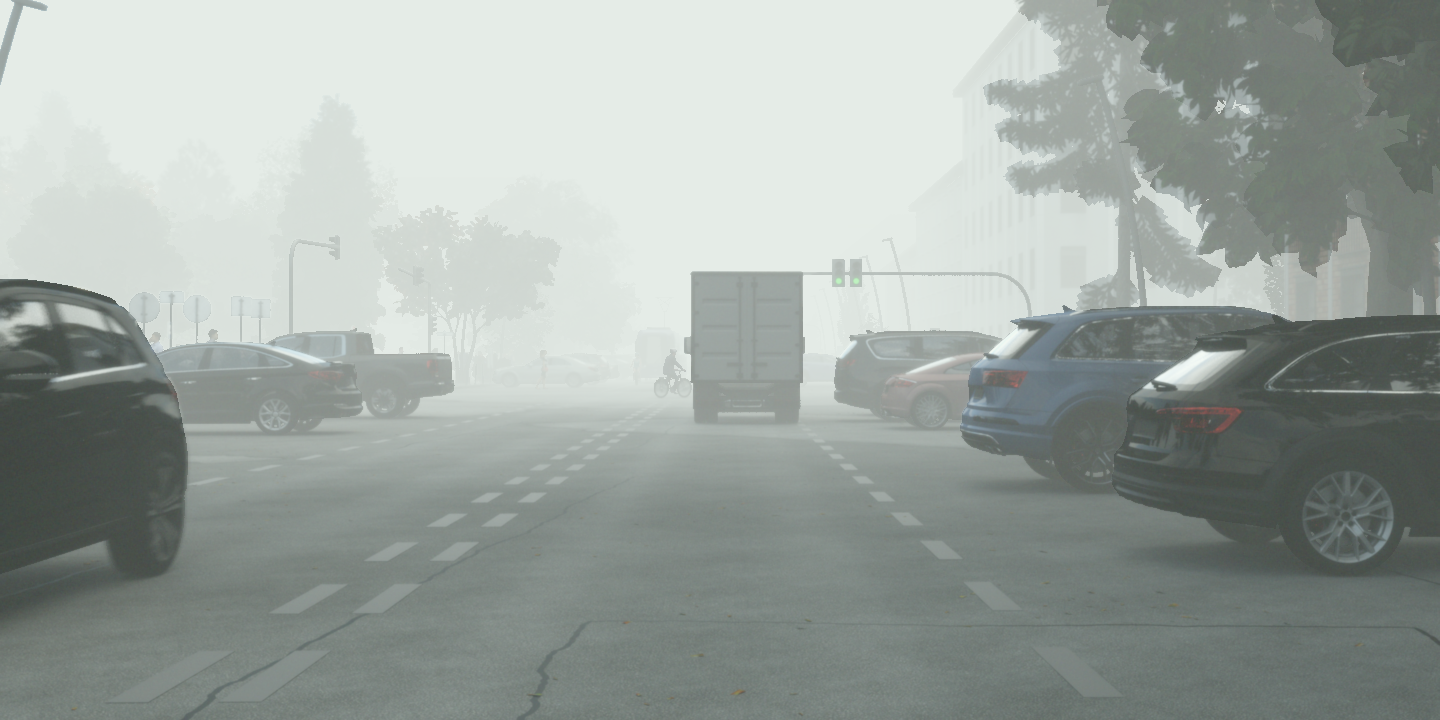} &
        \includegraphics[width=0.23\linewidth]{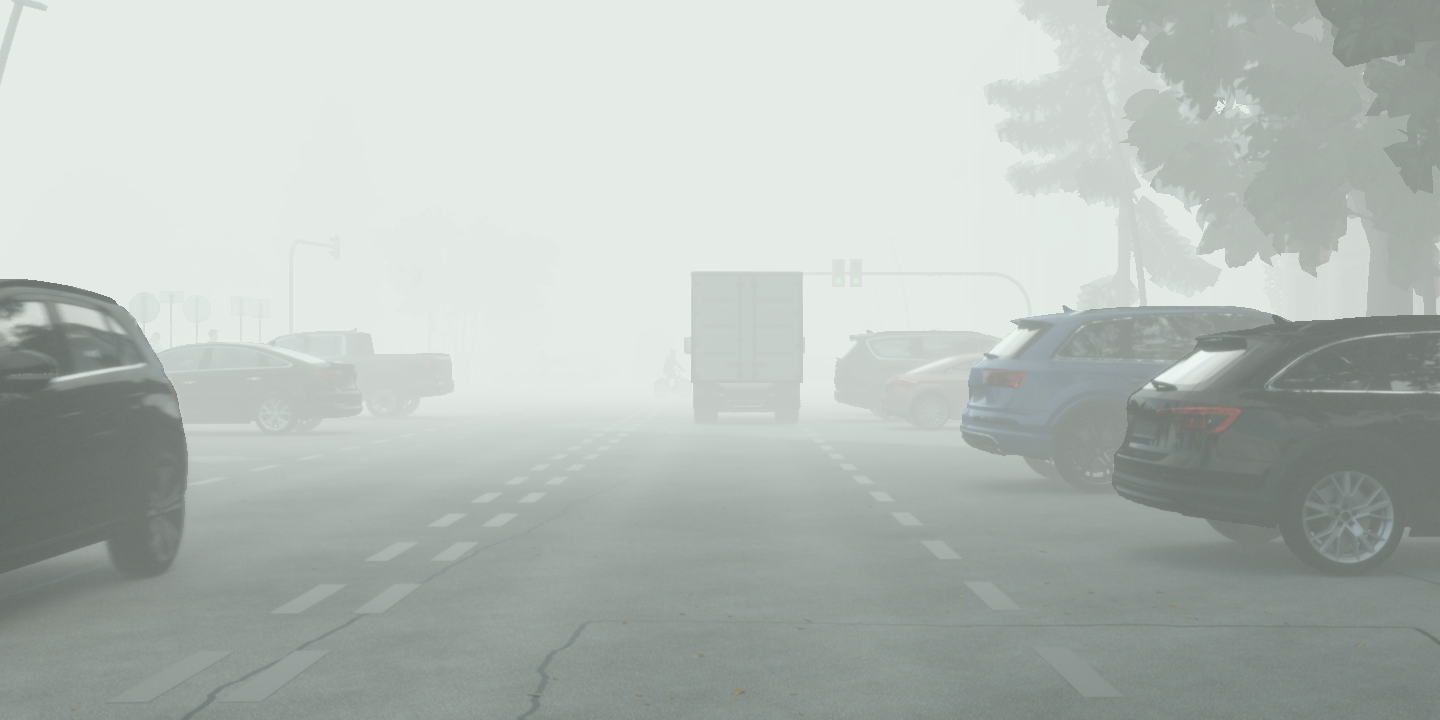} &
        \includegraphics[width=0.23\linewidth]{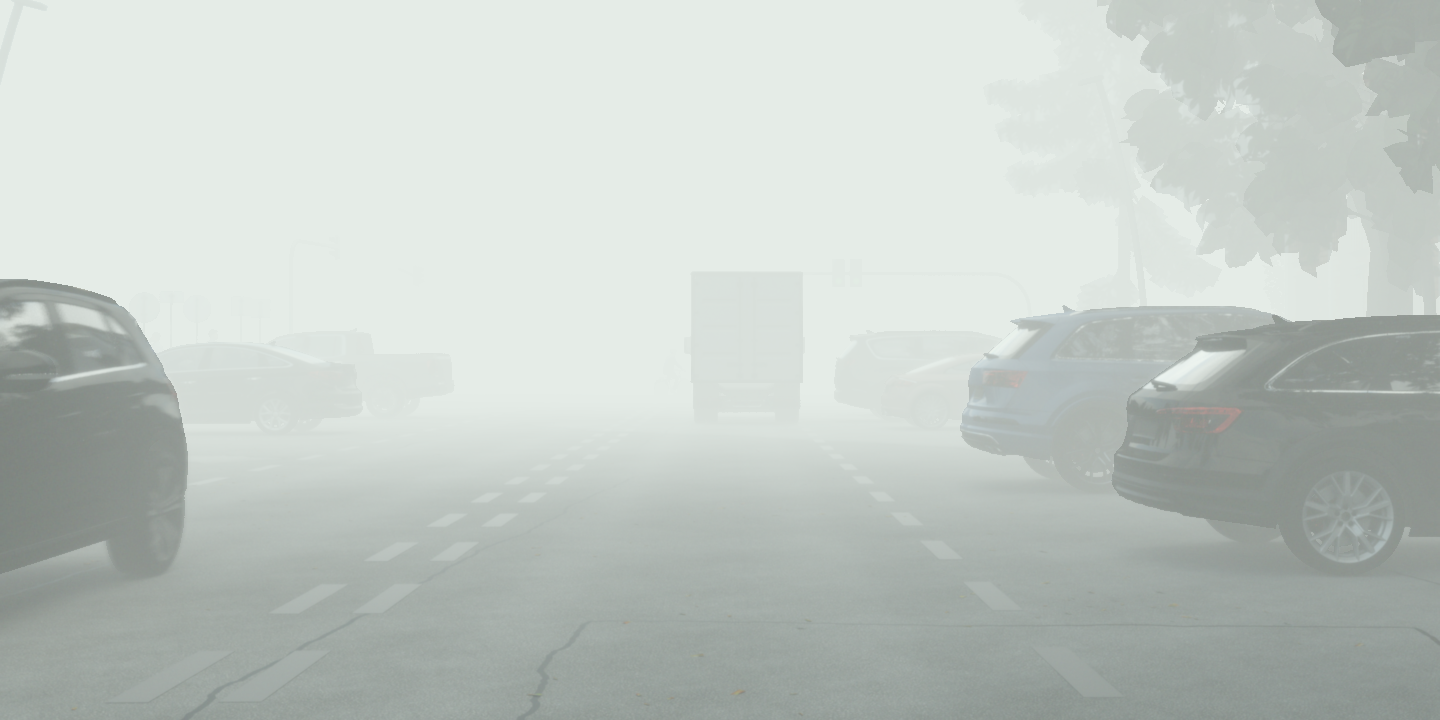} \\

        \includegraphics[width=0.23\linewidth]{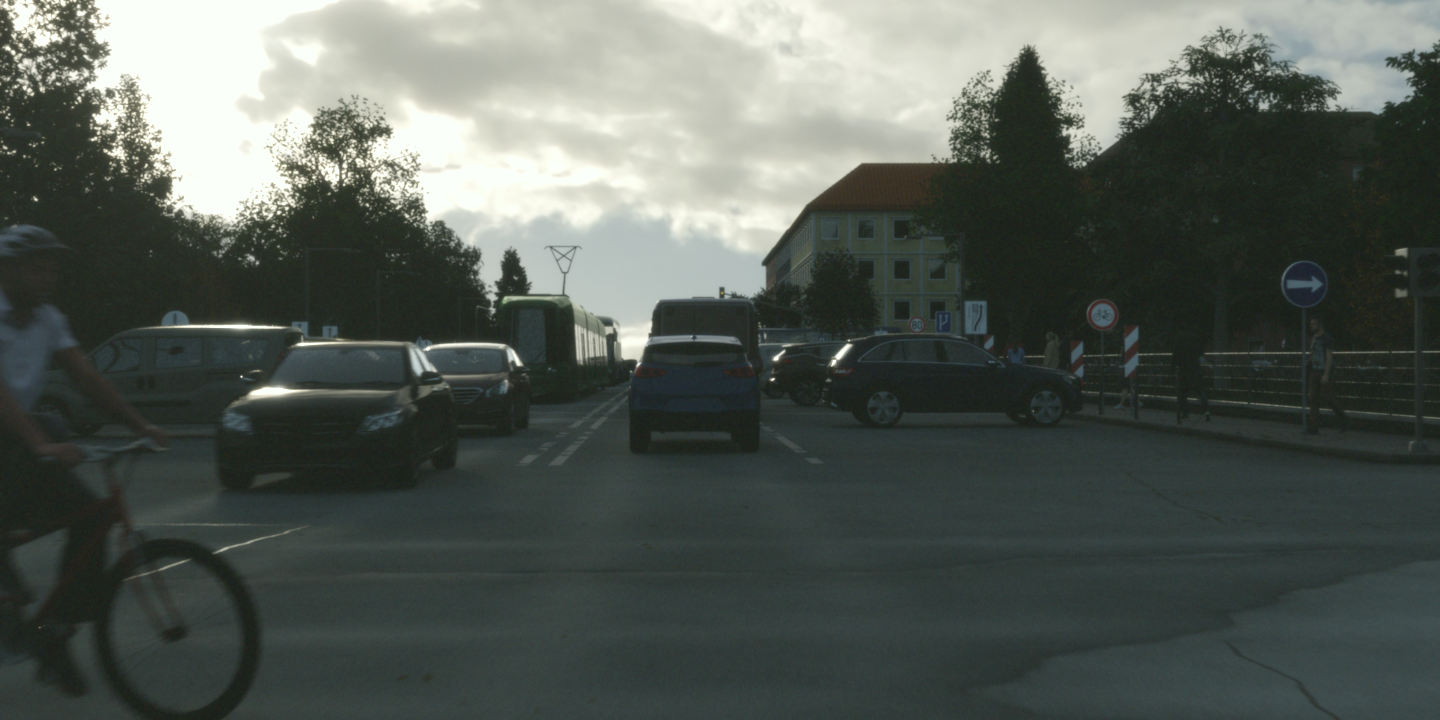} &
        \includegraphics[width=0.23\linewidth]{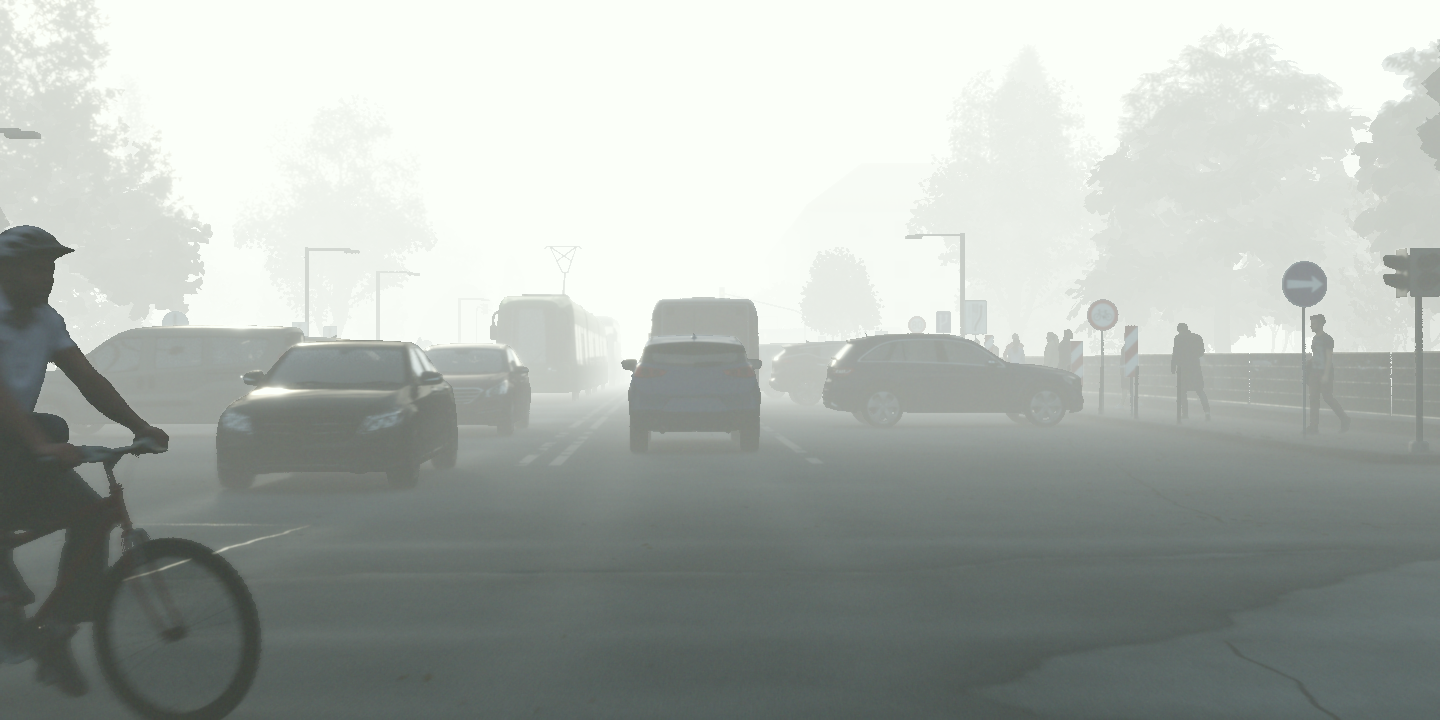} &
        \includegraphics[width=0.23\linewidth]{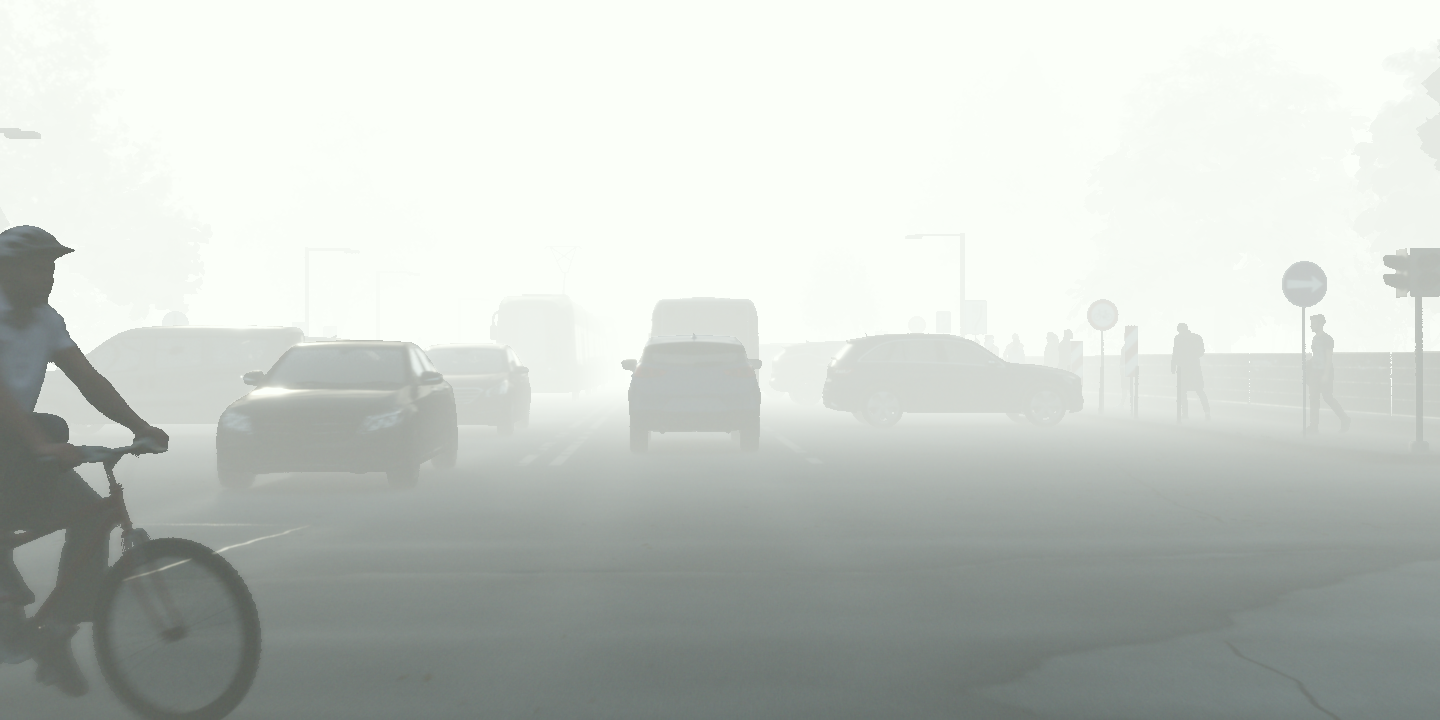} &
        \includegraphics[width=0.23\linewidth]{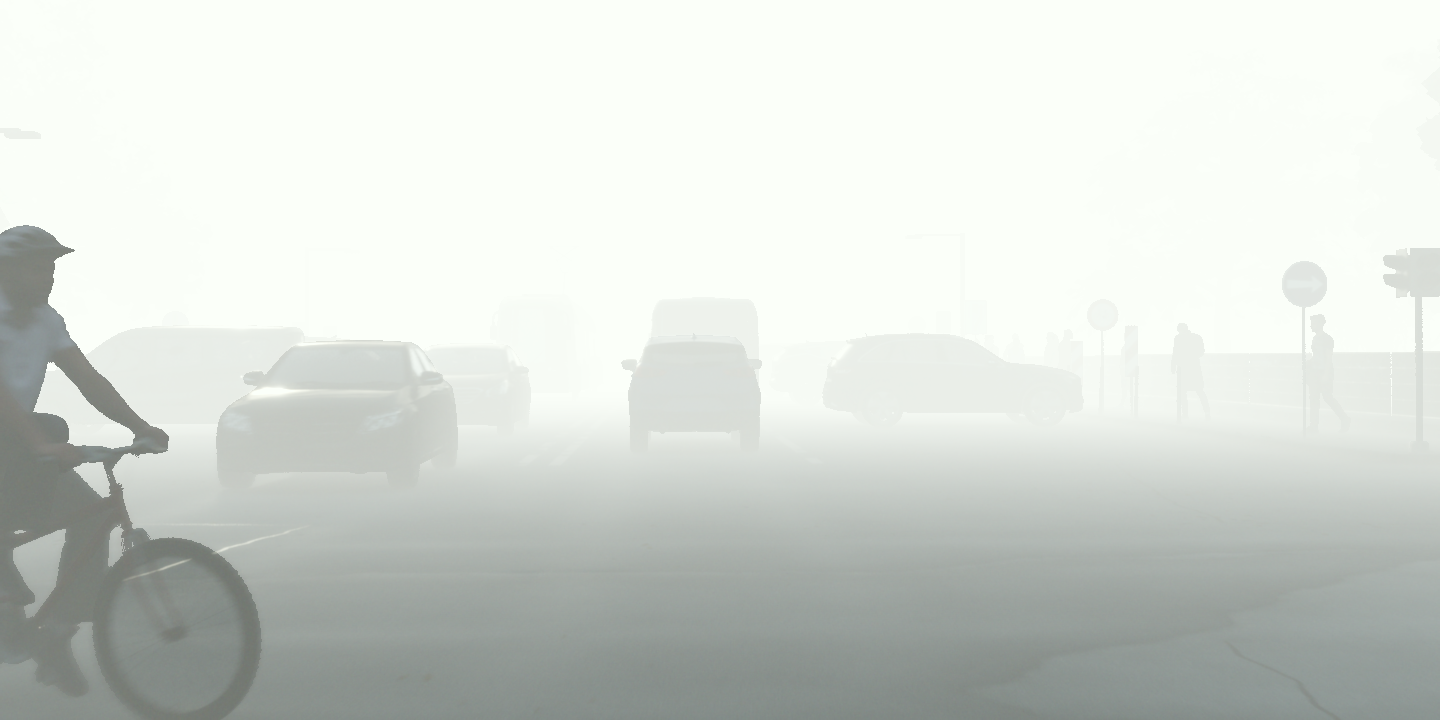} \\

         (a) clear & (b) \( \beta = 0.03 \) & (c) \( \beta = 0.06 \) & (d) \( \beta = 0.09 \) \\
    \end{tabular}
    \caption{Comparison of clear weather images from Synscapes \cite{wrenninge2018synscapes} against images from our adapted Foggy Synscapes for
\( \beta = 0.03 \), 0.06, 0.09.}
    \label{fig:dehazing_comparison_18_images}
\end{figure}
\subsection{ADAM-Dehaze Framework}

The overall pipeline is illustrated in Figure~\ref{fig:Flowchart}, showing the dataset creation, HDEN-based routing, and multi-branch adaptive dehazing architecture.

\subsubsection{Haze Density Estimation}

A lightweight \textit{Haze Density Estimation Network} (HDEN), denoted \( D_{\theta_D} \), estimates the fog intensity as:
\begin{equation}
d = D_{\theta_D}(P), \quad d \in [0,1],
\label{eq:score}
\end{equation}
where \( d \) is a continuous score indicating the haze level of image \( P \).

\subsubsection{Adaptive Branch Selection}
Given thresholds \( \alpha \) and \( \beta \), we route the input image \( P \) to one of three dehazing sub-networks:
\begin{equation}
J_{\mathrm{out}}(P) =
\begin{cases}
f_{\mathrm{Light}}(P), & d < \alpha,\\
f_{\mathrm{Medium}}(P), & \alpha \leq d \leq \beta,\\
f_{\mathrm{Complex}}(P), & d > \beta.
\end{cases}
\label{eq:branch}
\end{equation}

Let \( P_{LQ_i} \) denote the low-quality (hazy) input images categorized by estimated density level \( i \in \{1,2,3\} \), and \( P_{HQ_i} \) their corresponding restored outputs after passing through the appropriate CORUN branch:
\begin{itemize}
    \item \textbf{Light:} 2-stage unfolding for efficient processing of mildly foggy scenes.
    \item \textbf{Medium:} 4-stage unfolding to balance detail recovery and computational cost.
    \item \textbf{Complex:} 6-stage unfolding enhanced with transformer-based attention to restore visibility under dense fog.
\end{itemize}

\subsubsection{Adaptive Loss Function}

To balance fidelity and generalization, we propose a density-aware composite loss:
\begin{equation}
\begin{aligned}
\mathcal{L}_{\text{adaptive}} =\ & \gamma(d)\, \mathcal{L}_{\text{coh}} + (1 - \gamma(d))\, \mathcal{L}_{\text{contraRec}} + \mathcal{L}_{\text{dens}},\\
\gamma(d) &=
\begin{cases}
0.3, & d < \alpha, \\
0.6, & \alpha \le d \le \beta, \\
0.9, & d > \beta.
\end{cases}
\end{aligned}
\label{eq:adaptive_loss}
\end{equation}

\paragraph{Coherence Loss:} Enforces physical plausibility by minimizing the L1 distance between reconstructed and observed hazy images using the estimated transmission map \( T_{\text{out}} \):
\begin{equation}
\mathcal{L}_{\text{coh}} = \left\| J_{\text{out}} \odot T_{\text{out}} + (1 - T_{\text{out}}) - P \right\|_1.
\label{eq:coh}
\end{equation}

\paragraph{Perceptual Loss:} Encourages high-level perceptual similarity using pre-trained VGG-19 features:
\begin{equation}
\mathcal{L}_{\text{contraRec}} = \sum_{i} \tau_i \left\| \phi_i(J_{\text{gt}}) - \phi_i(J_{\text{out}}) \right\|_1,
\label{eq:perc}
\end{equation}
where \( \phi_i \) denotes the feature map from the \( i \)-th layer and \( \tau_i \) is its corresponding weight.

\paragraph{Density Loss:} Penalizes residual haze by minimizing the estimated haze score of the dehazed image:
\begin{equation}
\mathcal{L}_{\text{dens}} = \left\| D_{\theta_D}(J_{\text{out}}) \right\|_1.
\label{eq:dens}
\end{equation}

\subsubsection{Integration with Object Detection}

To validate utility in downstream tasks, we feed \( J_{\mathrm{out}} \) into a YOLOv8 object detector. Detection losses can be propagated back to fine-tune the dehazing branches for task-specific optimization, aligning restoration quality with detection performance.

\subsection{ADAM-Dehaze Algorithm}
\begin{algorithm}[H]
\caption{Adaptive Density-Aware Multi-Stage Image Dehazing (ADAM-Dehaze)}
\label{algo:adam-dehaze}
\begin{algorithmic}[1]
\REQUIRE Hazy image dataset $\{P_i\}_{i=1}^{N}$, Haze Density Estimation Network $D$, Adaptive Dehazing Models, Thresholds $\alpha$, $\beta$, Learning rate $\eta$, Epochs $E$
\ENSURE Dehazed images $\{J_i\}_{i=1}^{N}$
\STATE Initialize network parameters
\FOR{epoch $= 1$ to $E$}
    \FOR{each hazy image $P_i$}
        \STATE \textbf{Step 1:} Compute haze density score $d_i = D(P_i)$
        \STATE \textbf{Step 2:} Select appropriate model based on $d_i$
        \IF{$d_i < \alpha$}
            \STATE $J_i \gets$ Apply lightweight model on $P_i$
        \ELSIF{$\alpha \leq d_i \leq \beta$}
            \STATE $J_i \gets$ Apply medium-complexity model on $P_i$
        \ELSE
            \STATE $J_i \gets$ Apply high-complexity model on $P_i$
        \ENDIF
        \STATE \textbf{Step 3:} Set adaptive weight $\gamma$ based on density
        \IF{$d_i < \alpha$}
            \STATE $\gamma = 0.3$
        \ELSIF{$\alpha \leq d_i \leq \beta$}
            \STATE $\gamma = 0.6$
        \ELSE
            \STATE $\gamma = 0.9$
        \ENDIF
        \STATE \textbf{Step 4:} Compute adaptive loss $L$ using $\gamma$
        \STATE \textbf{Step 5:} Update network parameters using gradient descent
    \ENDFOR
\ENDFOR
\STATE \textbf{Inference:} Apply trained model to test images using Steps 1-2
\RETURN Dehazed images $\{J_i\}_{i=1}^{N}$
\end{algorithmic}
\end{algorithm}

\section{EXPERIMENTS AND RESULTS}

In this section, we present a comprehensive evaluation of the proposed ADAM-Dehaze framework. We first detail the implementation and dataset setup, then assess the haze‐density classifier. Next, we report quantitative and qualitative results for dehazing, followed by downstream object detection performance on both synthetic and real‐world datasets. Finally, we analyze computational complexity and perform ablation studies to validate each core component.

\subsection{Implementation Details}

The ADAM-Dehaze dehazing network is implemented in PyTorch and trained on an NVIDIA A100 GPU using mixed‐precision (APEX) to accelerate convergence and reduce memory usage. We employ the Adam optimizer with an initial learning rate of $2\times10^{-4}$, a weight decay of $10^{-4}$, and a batch size of 16 over 100 epochs. The learning rate is decayed by a factor of 0.1 at epochs 50 and 75. 

The object detector is YOLOv8n, pretrained on clear-weather images, then fine‐tuned on our synthetic FogIntensity‑25K dataset (see Table~\ref{tab:dataset_composition}) to align its feature distribution with dehazed inputs. We use a batch size of 32, a learning rate of $1\times10^{-4}$, and a mosaic augmentation strategy during fine‐tuning. All code, model weights, and the FogIntensity‑25K dataset will be publicly released for full reproducibility.





\subsection{Haze Density Classification}

Accurate haze classification is critical. We compare ten backbones on 5\,000 test images with ground‐truth fog labels. Table~\ref{tab:classifier_performance} shows that DenseNet121 achieves 99.80\% accuracy, 99.80\% precision/recall/F1, and a test loss of 0.0065, outperforming ResNet, VGG, and ConvNeXt variants. Its superior gradient flow and multi‐scale feature extraction explain its efficacy in distinguishing subtle haze levels.

\begin{table}[ht]
  \centering
  \setlength{\tabcolsep}{4.5pt}
  \caption{Haze density classification performance (5,000‐image test set).}
  \small
  \begin{tabular}{lcccc}
    \toprule
    \textbf{Model} & \textbf{Accuracy} & \textbf{Precision} & \textbf{Recall} & \textbf{Test Loss} \\
    \midrule
    \rowcolor[HTML]{F2F2F2} ConvNeXt-L   & 94.67\% & 94.67\% & 94.67\% & 0.1479 \\
    \rowcolor[HTML]{f6fbff} VGG16        & 96.67\% & 96.69\% & 96.67\% & 0.1491 \\
    \rowcolor[HTML]{eff6fc} ResNet50     & 99.00\% & 99.02\% & 99.00\% & 0.0348 \\
    \rowcolor[HTML]{DDEBF7} DenseNet121  & \textbf{99.80\%} & \textbf{99.80\%} & \textbf{99.80\%} & \textbf{0.0065} \\
    \bottomrule
  \end{tabular}
  \label{tab:classifier_performance}
\end{table}

\begin{table*}[htbp]
\centering
\caption{Detailed quantitative comparison of BRISQUE and NIMA scores across different haze levels (light, medium, and complex) using the Cityscapes dataset. Our method consistently achieves the lowest BRISQUE scores and highest NIMA scores across all haze densities, demonstrating superior performance in both perceptual and aesthetic image restoration quality under varying fog conditions.}

\resizebox{1.8\columnwidth}{!}{
\begin{tabular}{lcccccc}
\toprule
\textbf{Methods} & \multicolumn{3}{c}{\textbf{BRISQUE $\downarrow$}} & \multicolumn{3}{c}{\textbf{NIMA $\uparrow$}} \\
\cmidrule(lr){2-4} \cmidrule(lr){5-7}
 & \textbf{Light} & \textbf{Medium} & \textbf{Complex} & \textbf{Light} & \textbf{Medium} & \textbf{Complex} \\
\midrule
\rowcolor[HTML]{F2F2F2}
DAD  \cite{shao2020domain} & 76.8971 & 74.9572 & 76.0635 & 4.0896 & 4.0894 & 4.2280 \\
\rowcolor[HTML]{f6fbff}
MBDN \cite{dong2020multi} & 24.8952 & 28.0286 & 33.9047 & 4.8659 & 4.8606 & 4.9849 \\
\rowcolor[HTML]{eff6fc}
RIDCP \cite{wu2023ridcp} & 21.2840 & 18.2256 & 17.5208 & 4.8380 & 5.0017 & 5.1447 \\
\rowcolor[HTML]{e6f1fa}
PSD \cite{chen2021psd} & 16.5855 & 19.5653 & 26.7156 & 4.8842 & 4.7927 & 4.8176 \\
\rowcolor[HTML]{DDEBF7}
Ours  & \textbf{10.7785} & \textbf{9.8197} & \textbf{8.5499} & \textbf{4.9738} & \textbf{5.1294} & \textbf{5.2583} \\
\bottomrule
\end{tabular}}
\label{tab:briso_nima_comparison}
\end{table*}

\subsection{Quantitative Evaluation of Dehazing}


\subsubsection{Synthetic Full‑reference Metrics}
On Cityscapes where ground truths exist, we report BRISQUE/NIMA (Table~\ref{tab:briso_nima_comparison}) and SSIM/PSNR/LPIPS (Table~\ref{tab:ssim_psnr_cityscapes}) for light (\(\beta=0.03\)), medium (\(\beta=0.06\)), and complex (\(\beta=0.09\)) fog. ADAM‑Dehaze consistently achieves the lowest distortion scores and highest perceptual/structural metrics, e.g., SSIM of 0.9188 and PSNR of 23.95 dB under light fog, confirming its ability to recover both global appearance and fine details.

\subsubsection{Real-World Non‑reference Metrics}
On the RTTS dataset, we employ FADE for haze density, BRISQUE for spatial quality, and NIMA for aesthetic score. Table~\ref{tab:combined_rtts_results} compares ADAM‑Dehaze against five state‑of‑the‑art methods. Our approach obtains the lowest FADE (0.824), indicating the most effective haze removal, the lowest BRISQUE (11.956) for best spatial fidelity, and the highest NIMA (5.342), reflecting superior perceptual appeal. These improvements exceed RIDCP’s performance by 12.7\% in FADE and 30.8\% in BRISQUE, demonstrating the efficacy of intensity‐aware dehazing in real‐world conditions.



\begin{figure}[htbp]
    \centering
    \setlength{\tabcolsep}{1pt} 
    \resizebox{\columnwidth}{!}{
    \begin{tabular}{cccc}

        \includegraphics[width=0.25\linewidth]{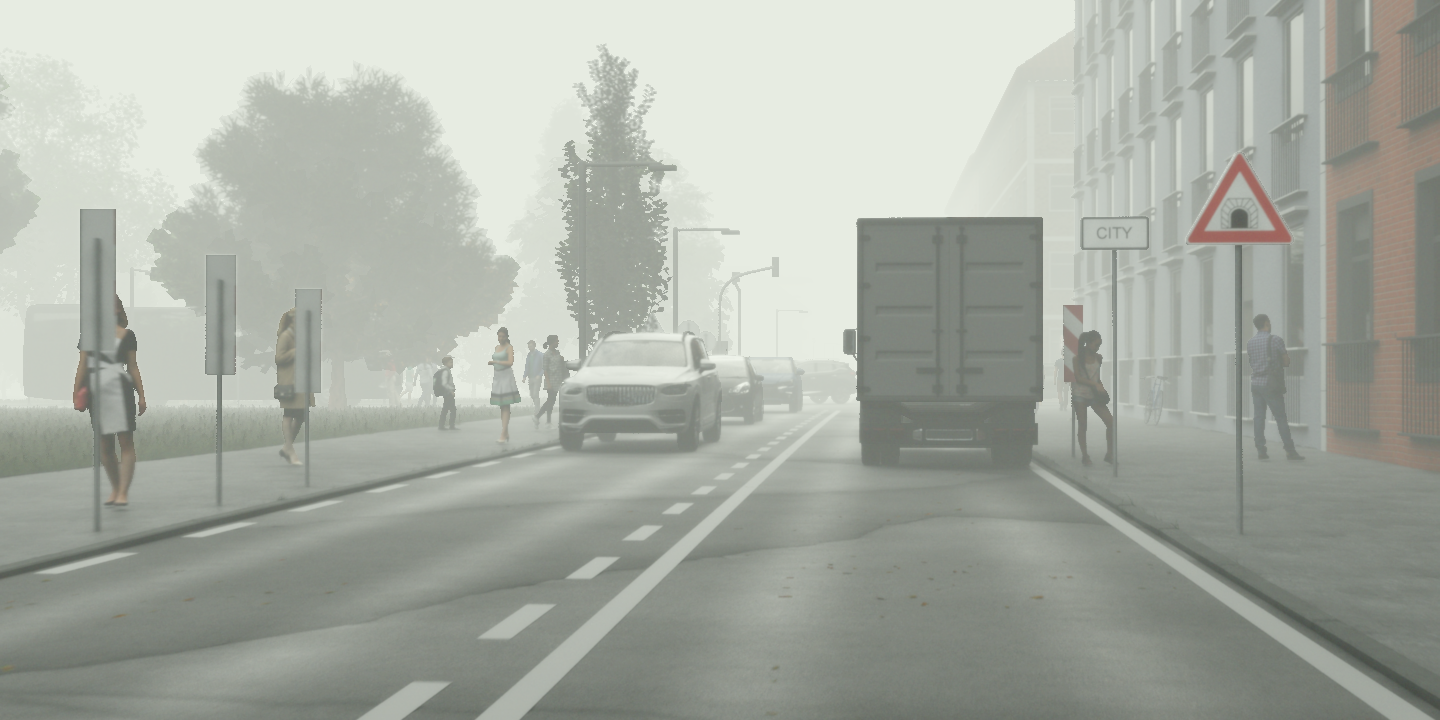} &
        \includegraphics[width=0.25\linewidth]{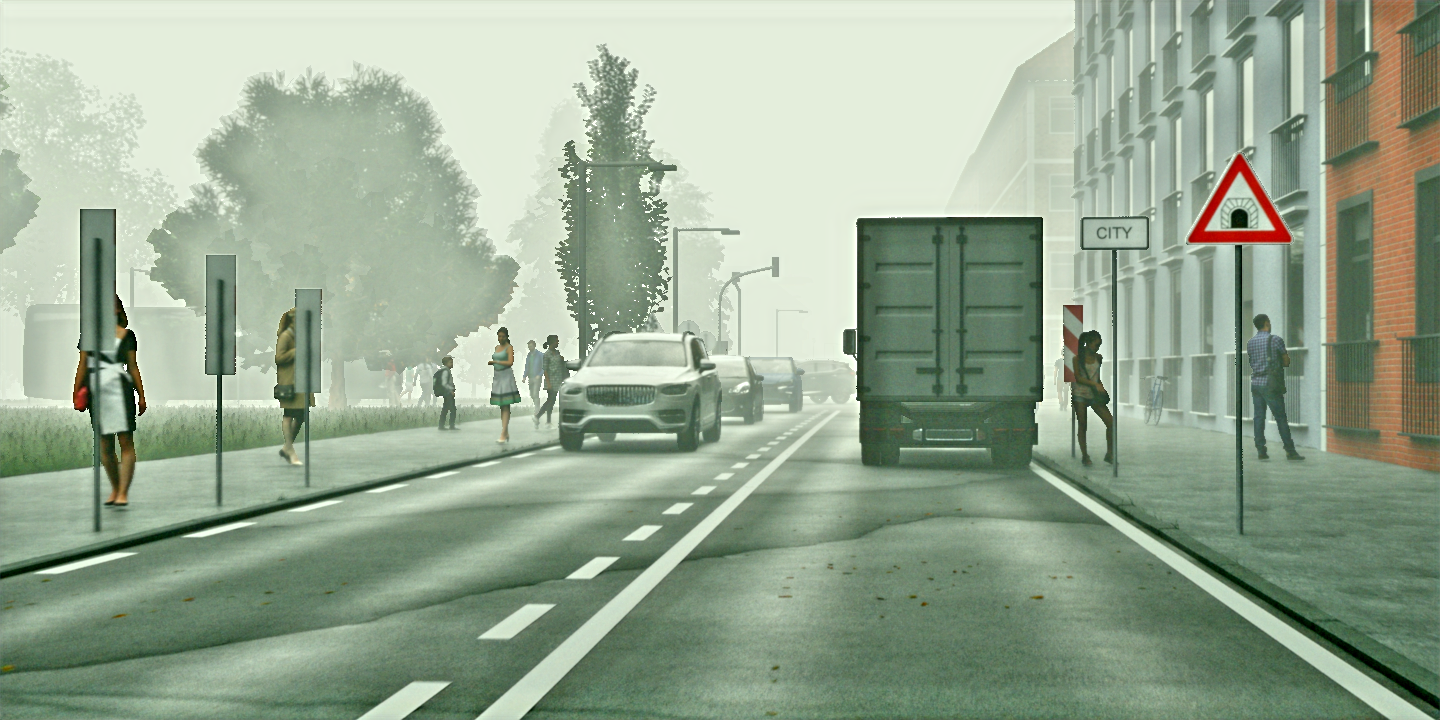} &
        \includegraphics[width=0.25\linewidth]{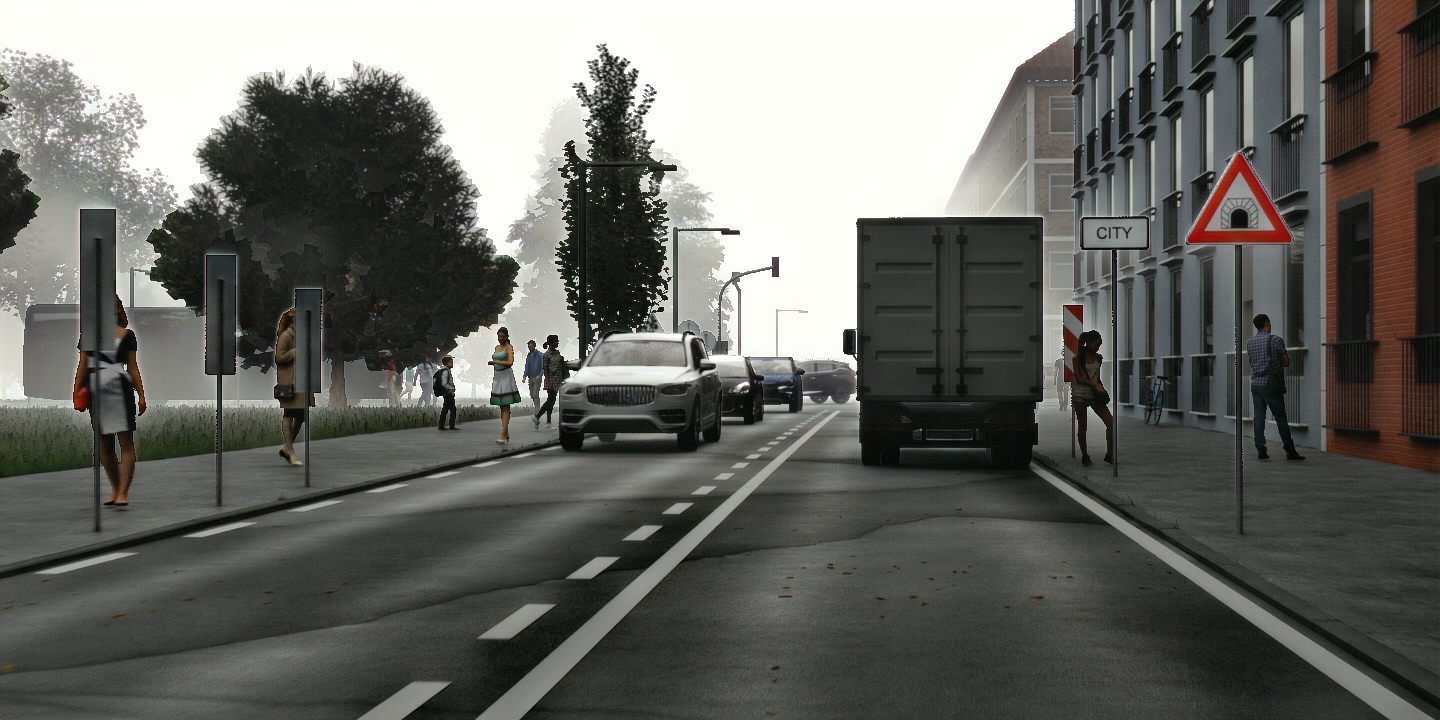} &
        \includegraphics[width=0.25\linewidth]{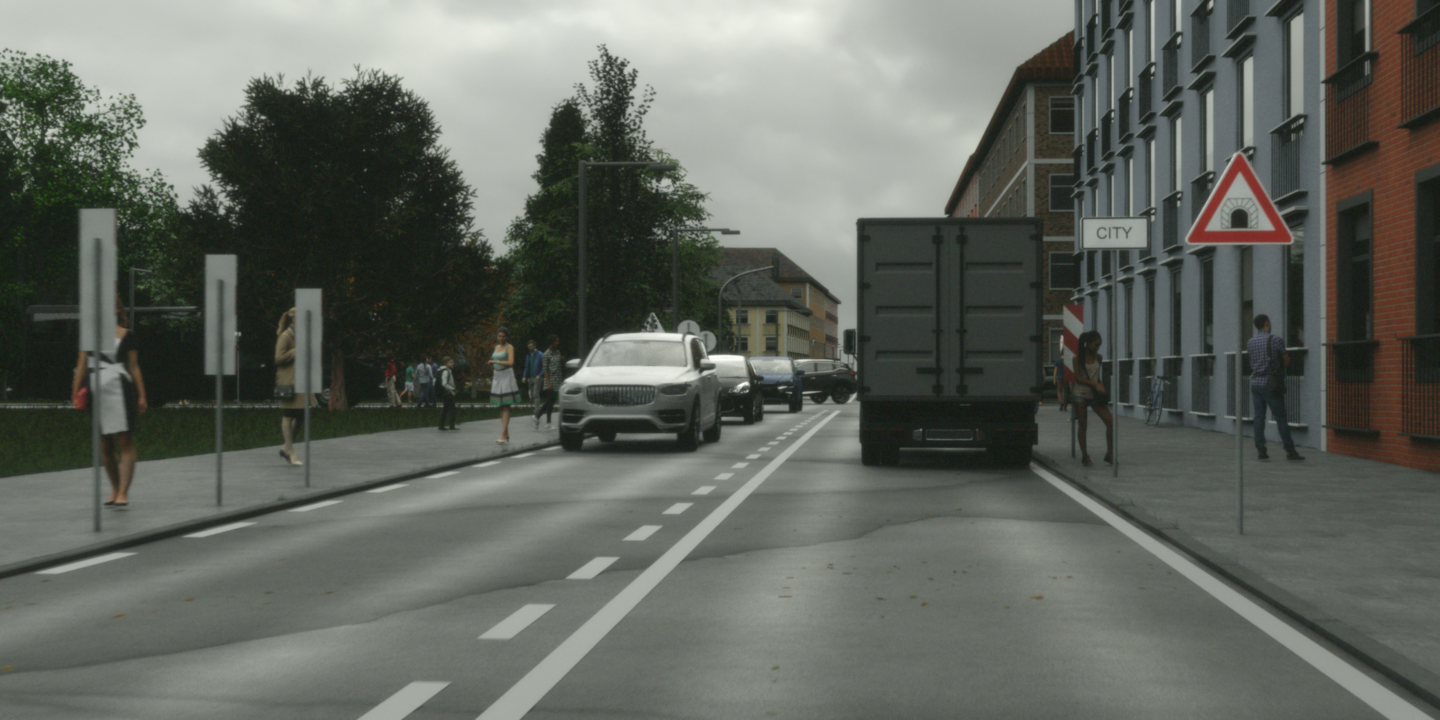} \\
        \\

        \includegraphics[width=0.25\linewidth]{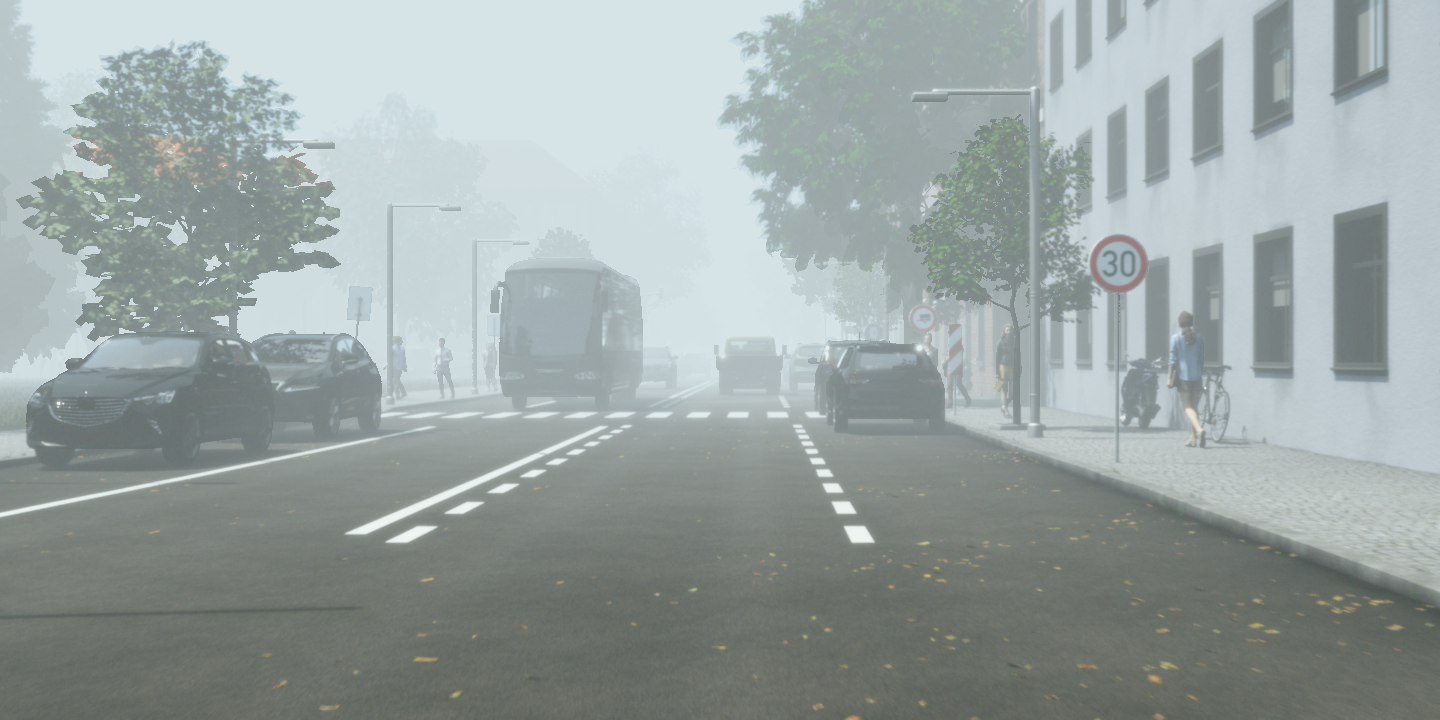} &
        \includegraphics[width=0.25\linewidth]{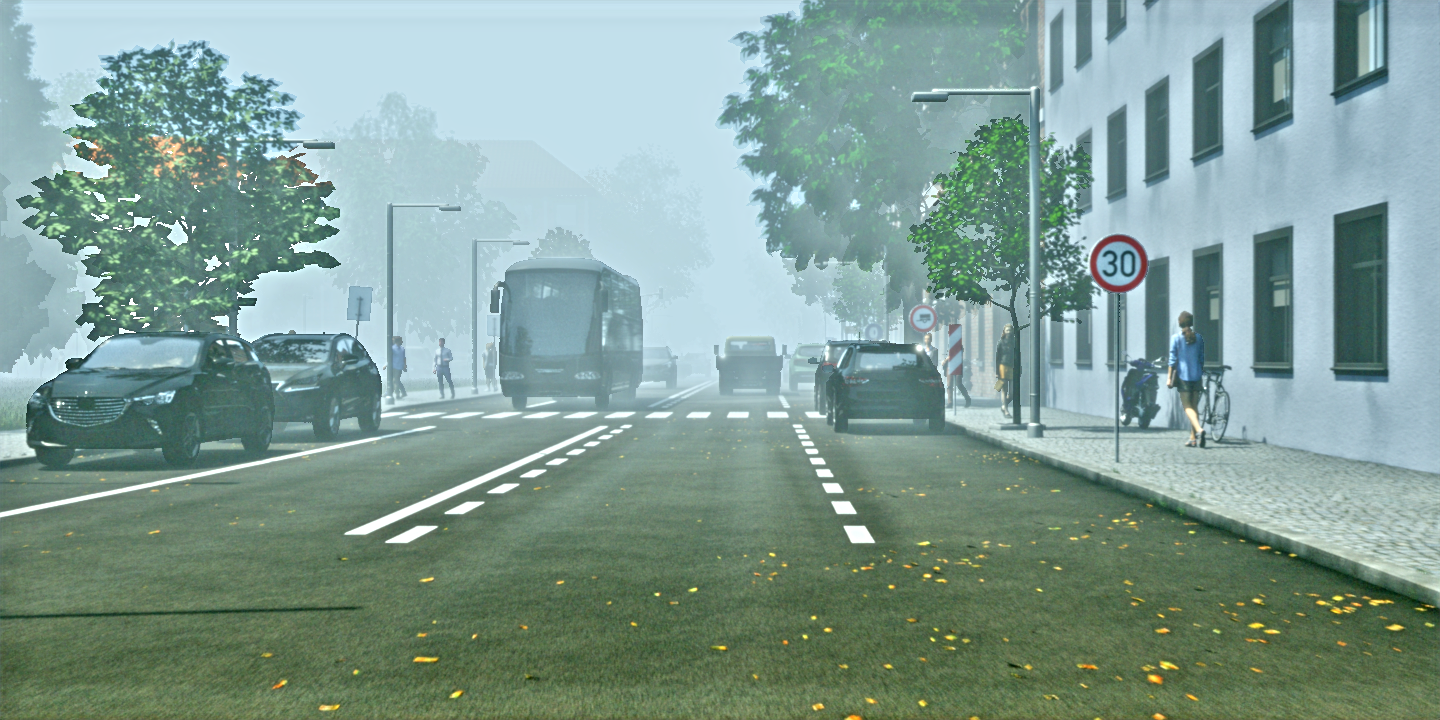} &
        \includegraphics[width=0.25\linewidth]{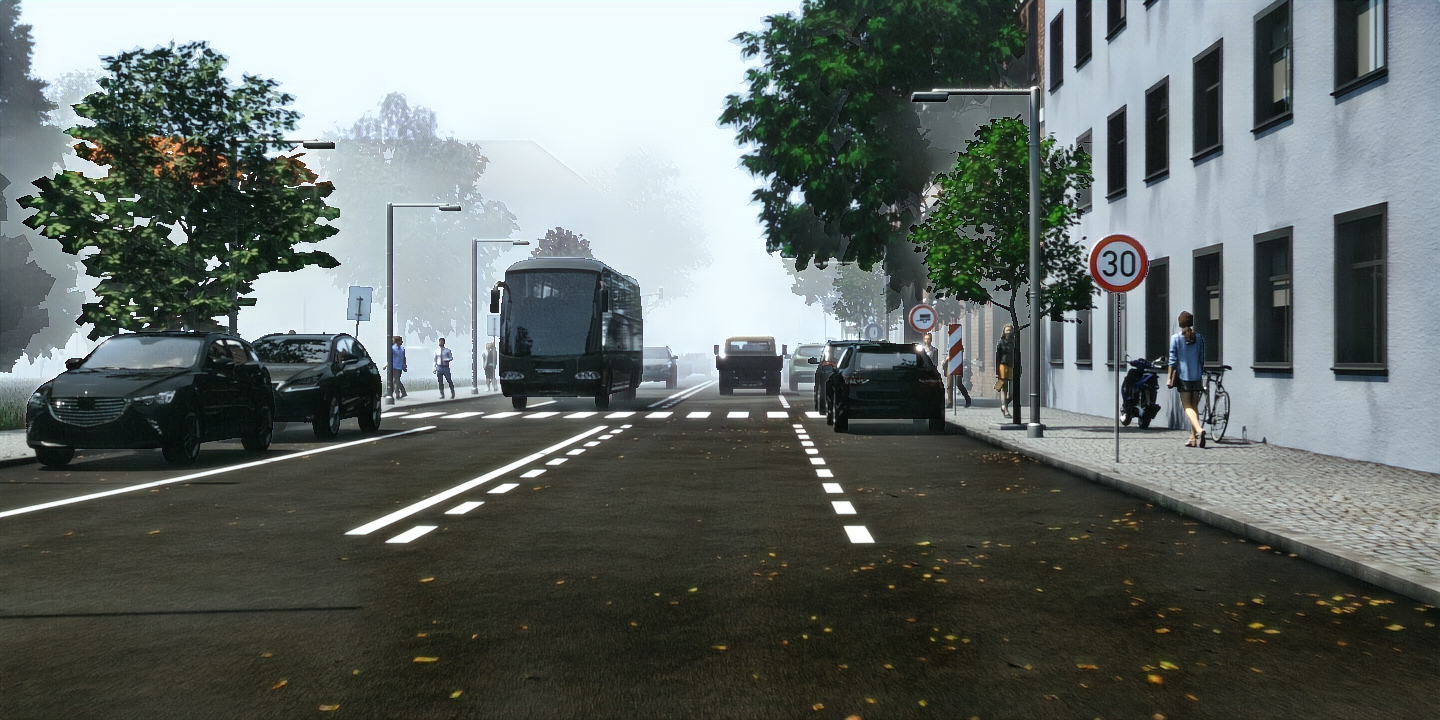} &
        \includegraphics[width=0.25\linewidth]{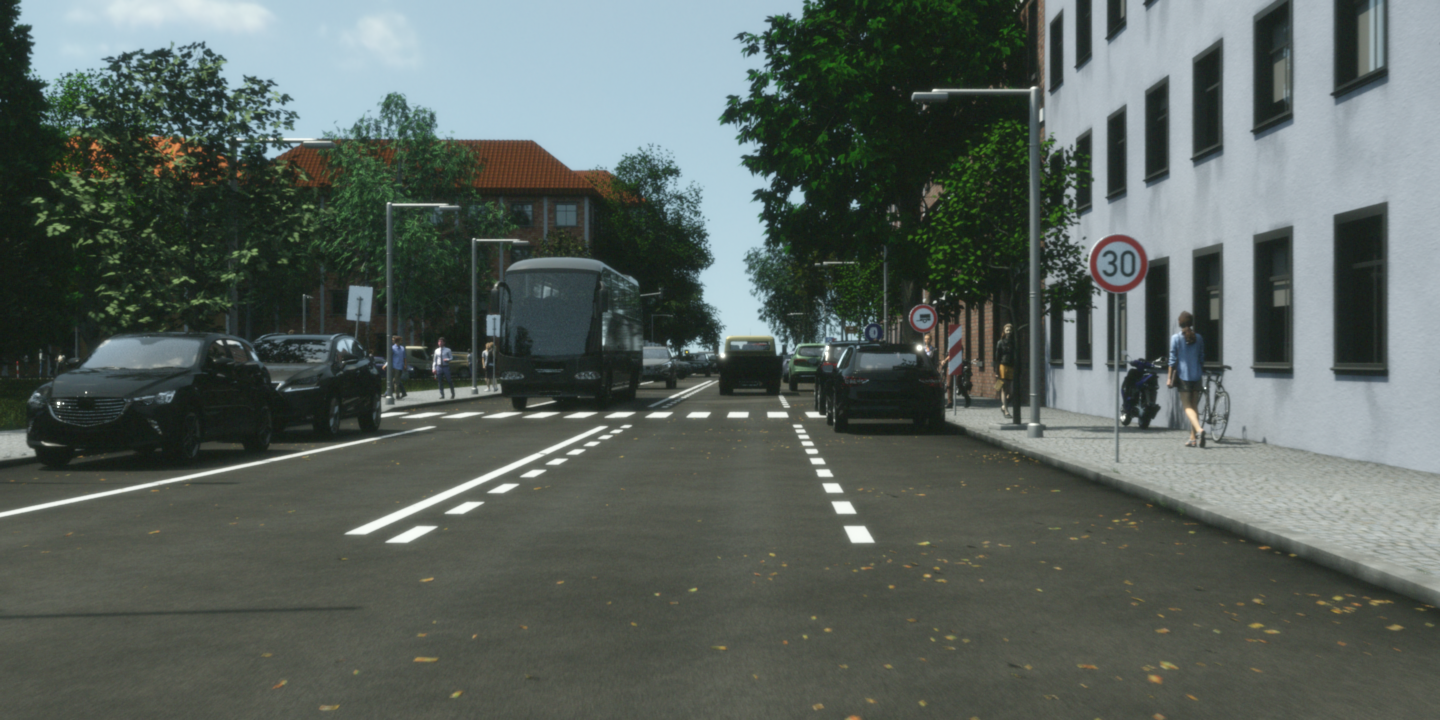} \\
         \\

        Hazy  & PSD \cite{chen2021psd} & Ours & GT \\
    \end{tabular}}
    \caption{Qualitative comparison of dehazing performance with PSD \cite{chen2021psd} on the Cityscapes dataset under light fog conditions ($\beta = 0.03$). Our method consistently produces clearer and more natural, closely resembling the GT images, particularly in restoring structural details and preserving color fidelity across diverse urban scenes.}
    \label{fig:dehazing_comparison_18_images}
\end{figure}


\begin{table}
\centering
\setlength{\tabcolsep}{12pt}
\caption{SSIM, PSNR and LPIPS evaluation for our method on the Cityscapes dataset under different fog levels. Higher SSIM and PSNR while lower LPIPS values indicate better image quality.}
\resizebox{\columnwidth}{!}{
\begin{tabular}{lccc}
\toprule
\textbf{Fog Level}      & \textbf{SSIM$\uparrow$} & \textbf{PSNR$\uparrow$} & \textbf{LPIPS$\downarrow$} \\
\midrule
\rowcolor[HTML]{eff6fc}
\textbf{Light}     & 0.9188       & 23.95       & 0.0585 \\
\rowcolor[HTML]{eff6fc}
\textbf{Medium}    & 0.8761       & 21.78       & 0.0929 \\
\rowcolor[HTML]{eff6fc}
\textbf{Complex}   & 0.8060       & 19.39       & 0.1456 \\
\bottomrule
\end{tabular}}
\label{tab:ssim_psnr_cityscapes}
\end{table}

\subsection{Qualitative Results}
 In figure ~\ref{fig:dehazing_comparison_18_images}, we observe that the dehazed outputs produced by our method exhibit noticeably higher clarity, better structural detail, and closer color fidelity relative to the input hazy images and even when compared to the outputs of the PSD and other methods. Collectively, these quantitative metrics establish that the proposed ADAM-Dehaze framework significantly outperforms current state-of-the-art methods in restoring image quality under varying fog conditions.


\begin{figure}[htbp]
    \centering
    \setlength{\tabcolsep}{1pt}
    \begin{tabular}{ccccc}
                
        \includegraphics[width=0.25\linewidth]{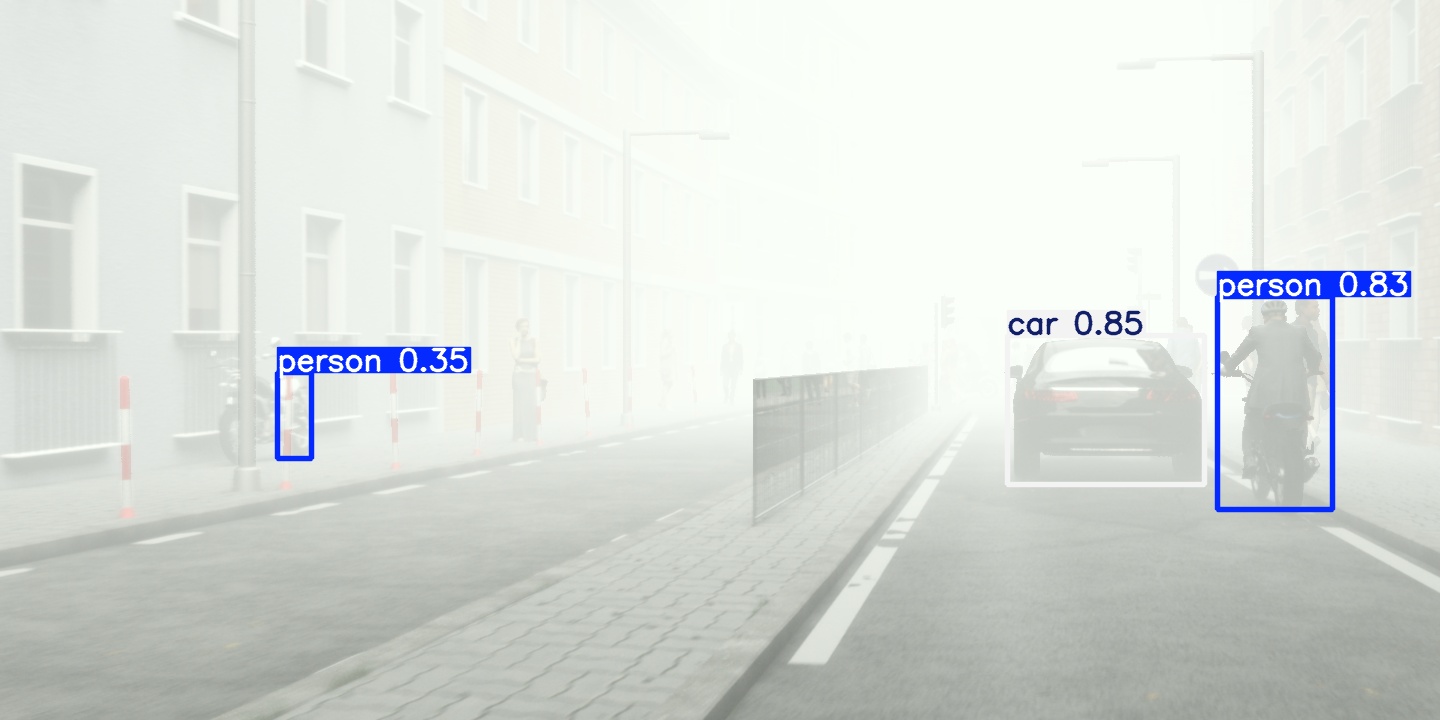} &
        \includegraphics[width=0.25\linewidth]
        {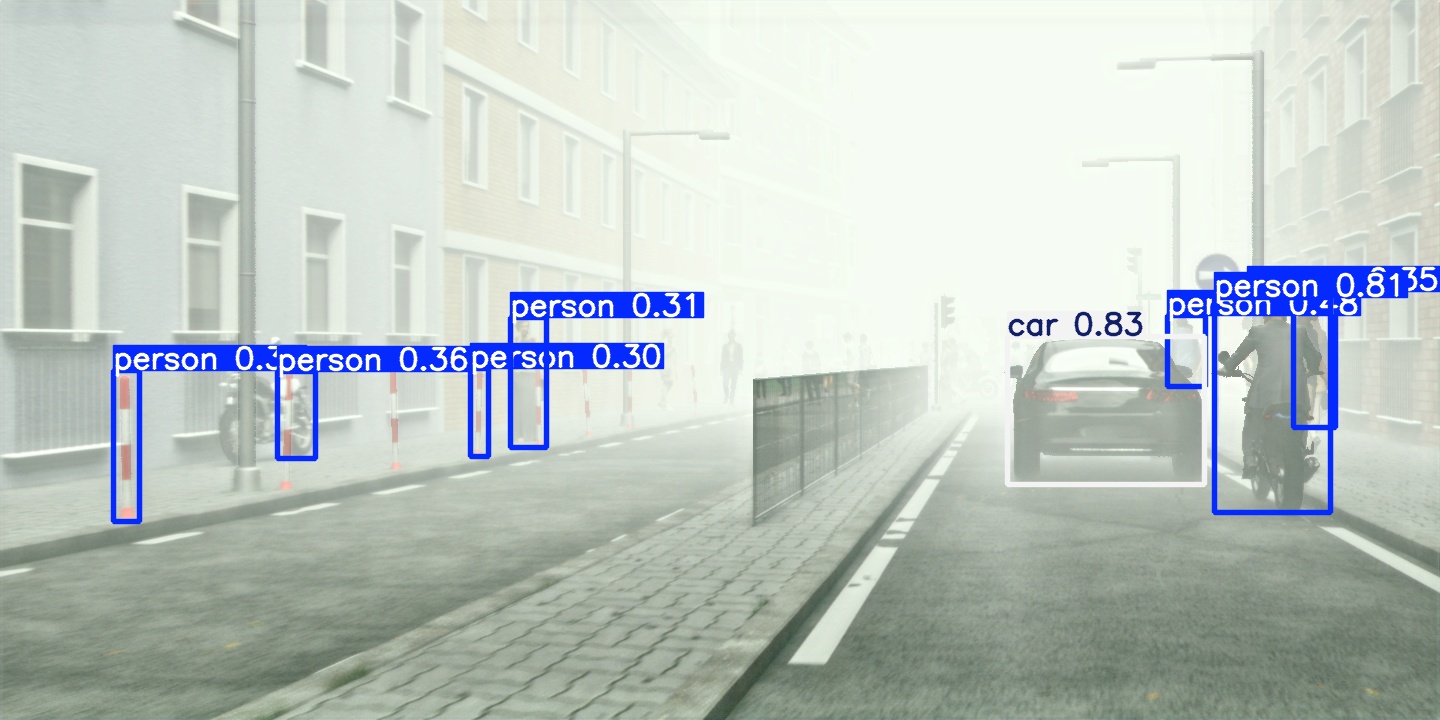} &
        \includegraphics[width=0.25\linewidth]{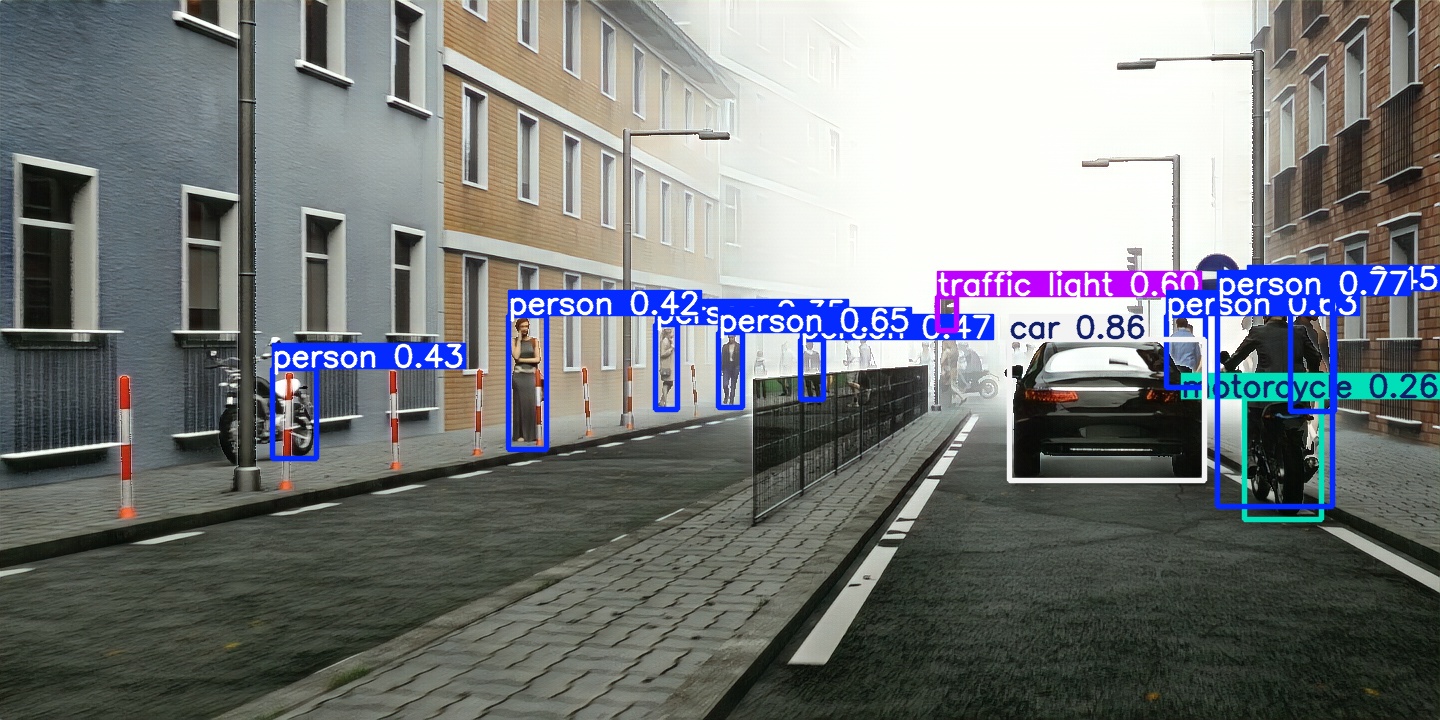} &
        \includegraphics[width=0.25\linewidth]{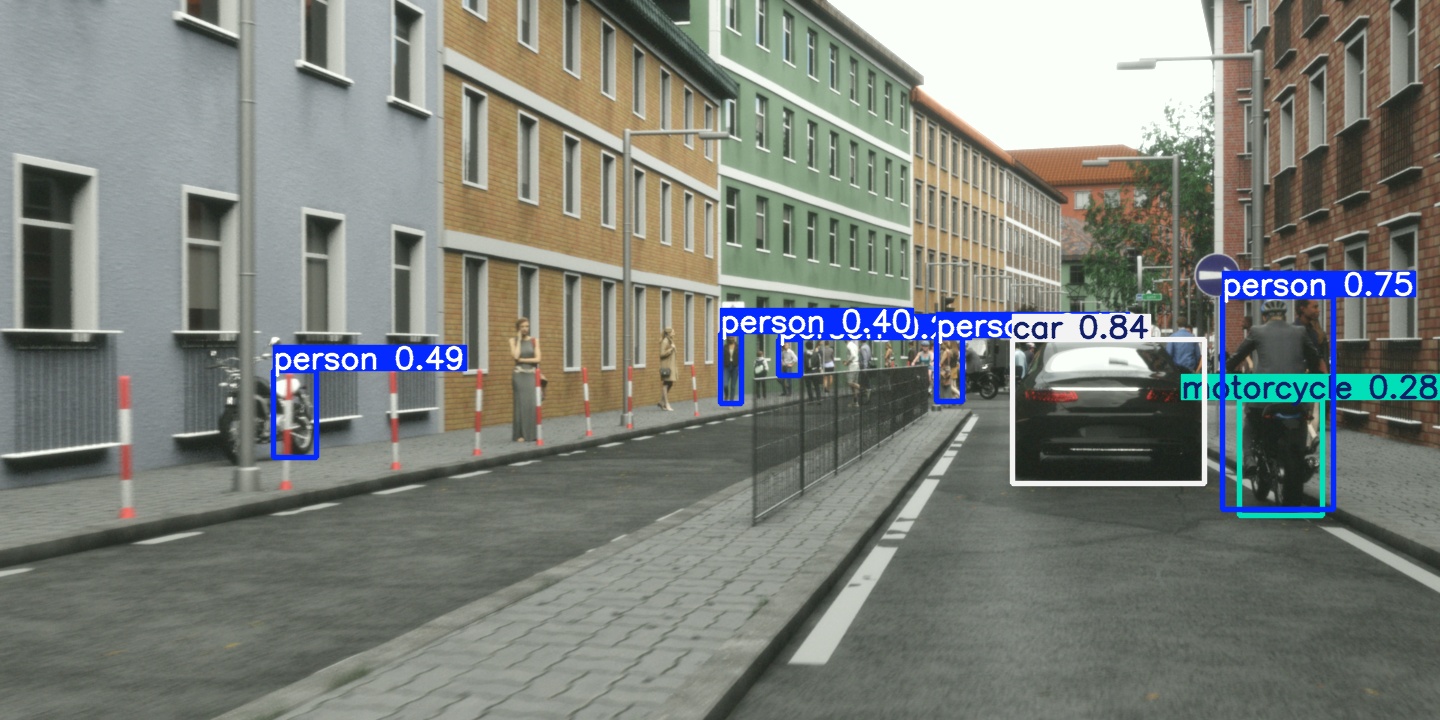} \\

        \includegraphics[width=0.25\linewidth]{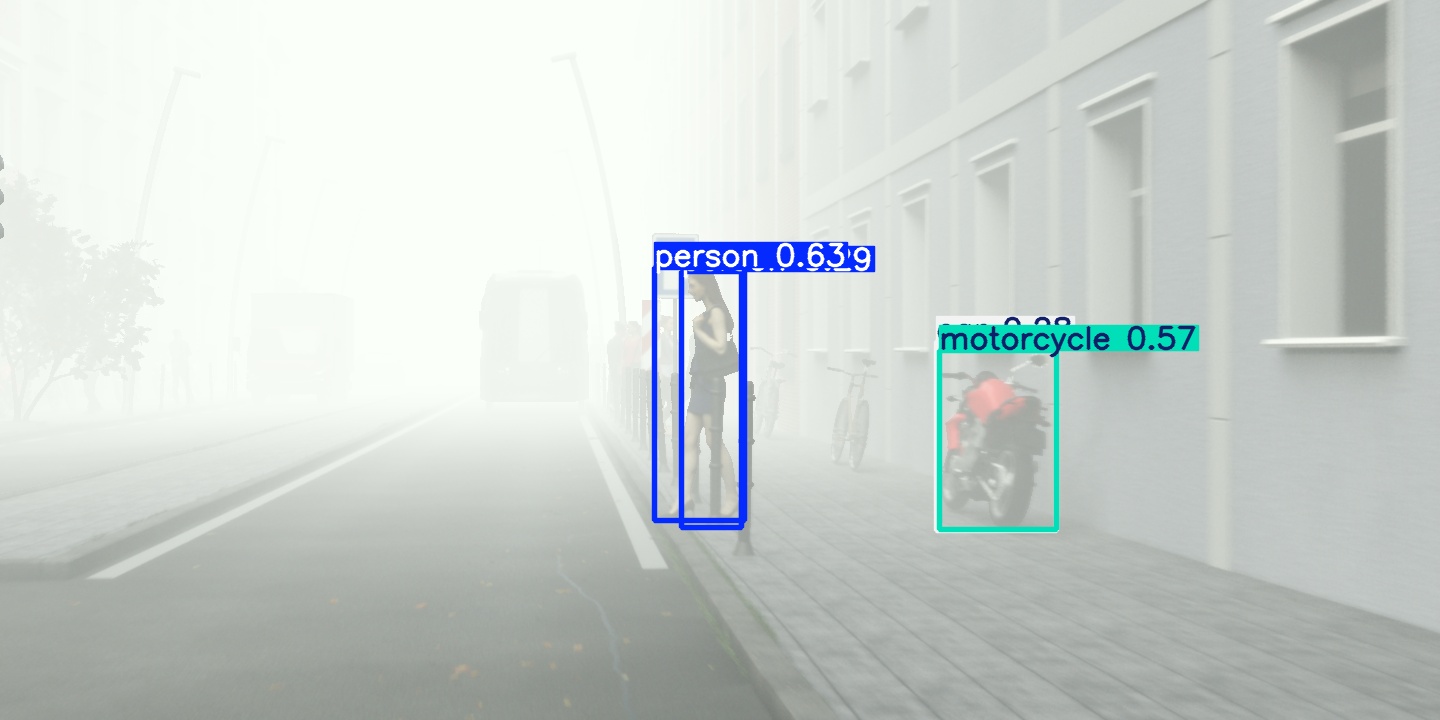} &
        \includegraphics[width=0.25\linewidth]
        {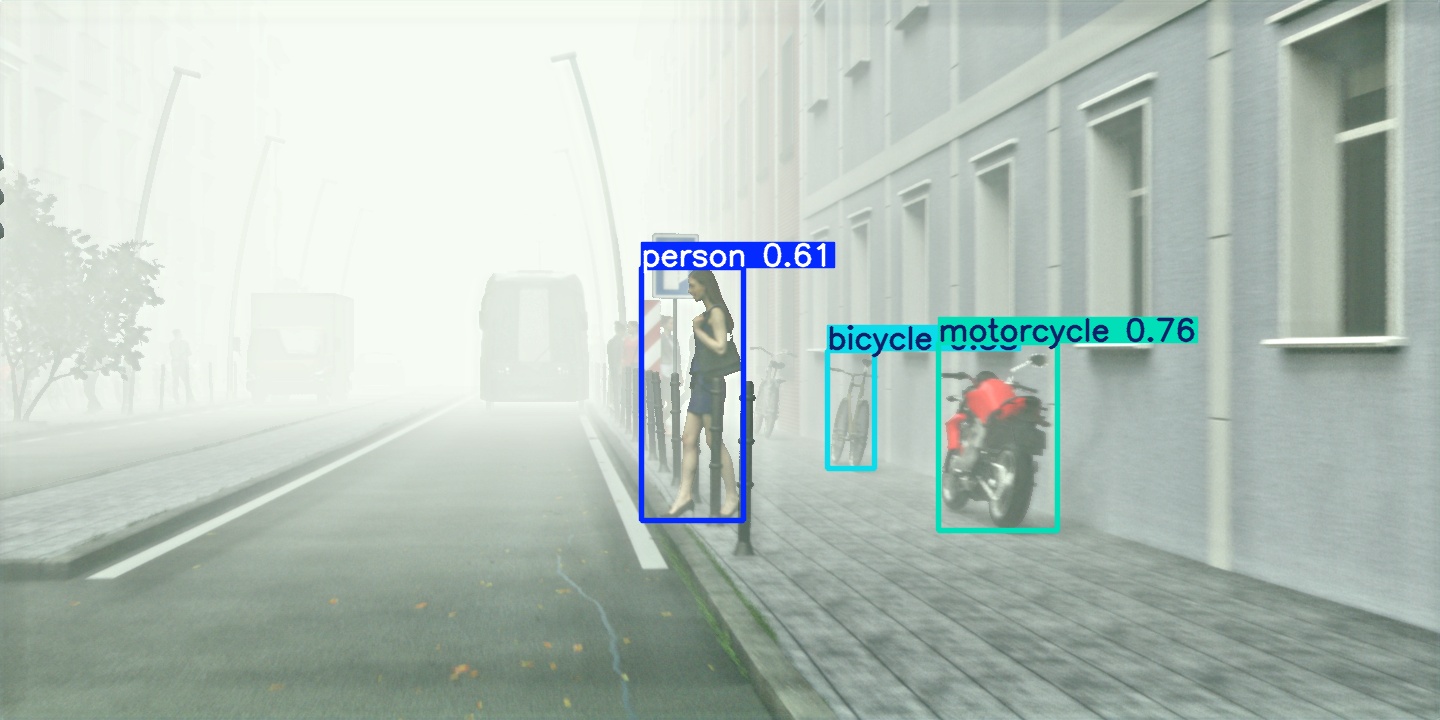} &
        \includegraphics[width=0.25\linewidth]{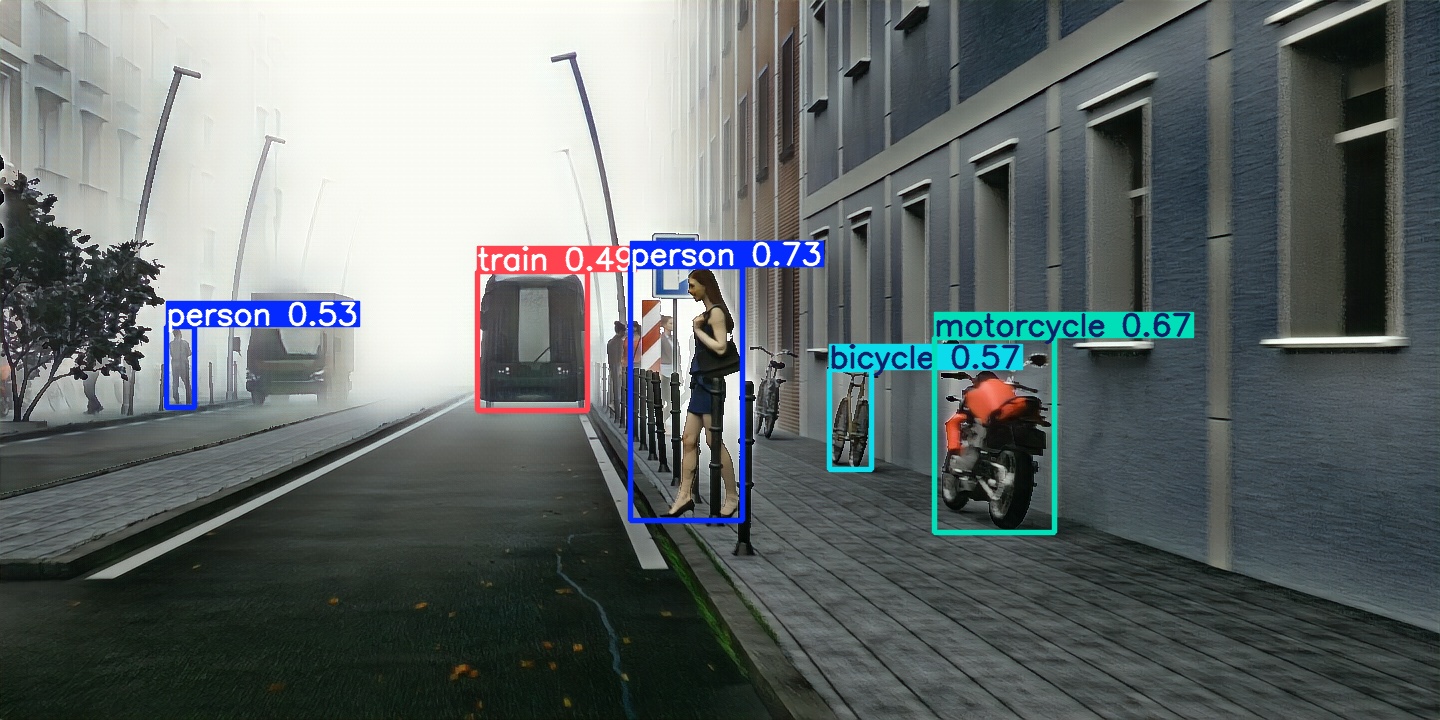} &
        \includegraphics[width=0.25\linewidth]{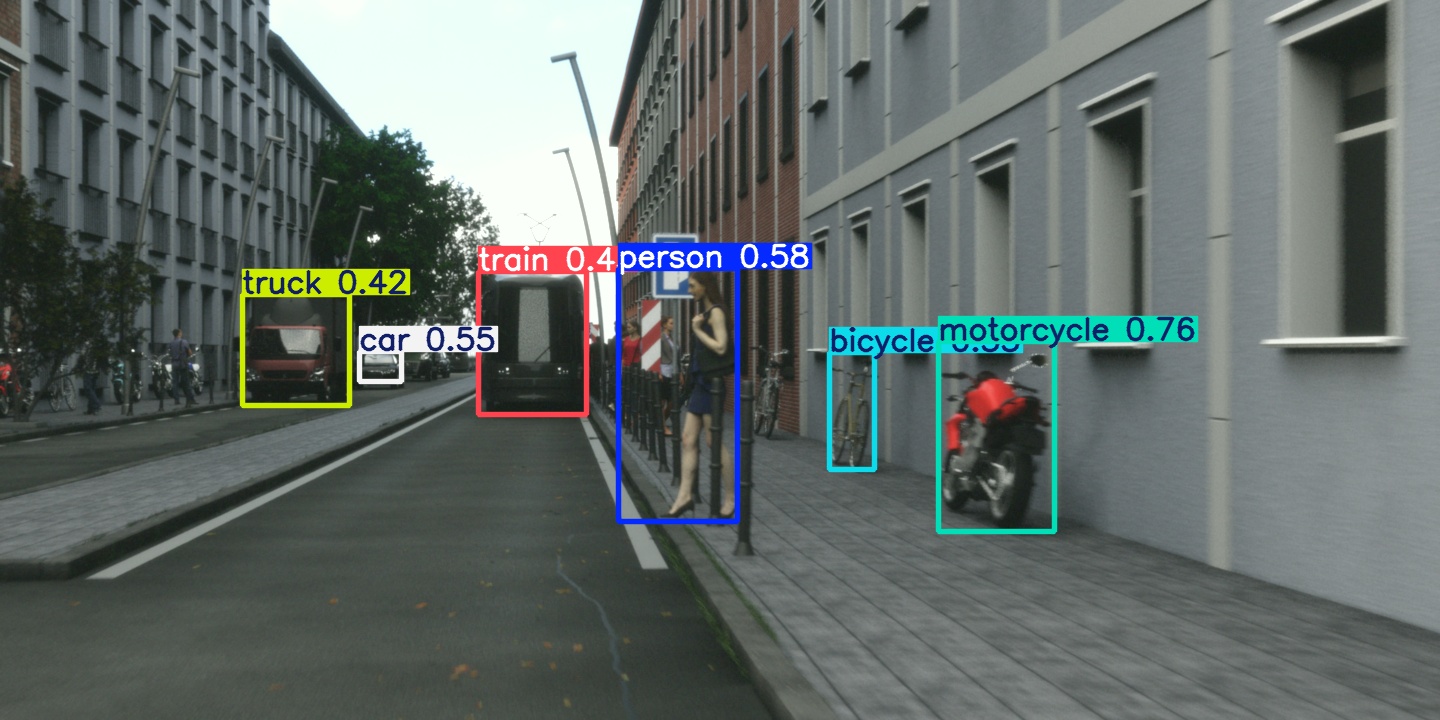} \\

        Hazy &  PSD & Ours & GT \\
    \end{tabular}
    \caption{Object detection under complex haze using PSD and our method ($\beta = 0.09$). Qualitative comparison with other dehazing methods is shown in appendix.}
    \label{fig:object_detection_heavy}
\end{figure}

\begin{table*}[htbp]
\centering
\caption{Quantitative results on the RTTS dataset under light fog conditions. FADE is used as a non-reference metric tailored for real-world hazy images, while BRISQUE and NIMA assess overall image quality. Additionally, object detection performance (mAP, \%) is reported for various categories. Our method, leveraging the CORUN-Light branch, outperforms baseline methods across both aspects.}
\resizebox{2.0\columnwidth}{!}{
\begin{tabular}{lccc|cccccc}
\toprule
\multirow{2}{*}{\textbf{Methods}} & \multicolumn{3}{c|}{\textbf{Image Quality (RTTS)}} & \multicolumn{6}{c}{\textbf{Object Detection (mAP, \%)}} \\ 
\cmidrule(lr){2-4} \cmidrule(lr){5-10}
 & \textbf{FADE $\downarrow$} & \textbf{BRISQUE $\downarrow$} & \textbf{NIMA $\uparrow$} & \textbf{Bicycle} & \textbf{Bus} & \textbf{Car} & \textbf{Motor} & \textbf{Person} & \textbf{Mean} \\ 
\midrule
\rowcolor[HTML]{F2F2F2}
MBDN \cite{dong2020multi} & 1.363 & 27.672 & 4.529 & 0.54 & 0.27 & 0.63 & 0.43 & 0.75 & 0.52 \\
\rowcolor[HTML]{F6FBFF}
DADv1 \cite{shao2020domain} & 1.130 & 32.241 & 4.312 & 0.52 & 0.29 & 0.65 & 0.38 & 0.74 & 0.51 \\
\rowcolor[HTML]{EFF6FC}
DADv2 \cite{shao2020domain} & 1.358 & 33.210 & 4.484 & 0.54 & 0.28 & 0.64 & 0.42 & 0.75 & 0.52 \\
\rowcolor[HTML]{E6F1FA}
PSD \cite{chen2021psd} & 0.920 & 27.713 & 4.598 & 0.52 & 0.25 & 0.63 & 0.42 & 0.74 & 0.51 \\
\rowcolor[HTML]{D9E9F6}
RIDCP \cite{wu2023ridcp} & 0.944 & 17.293 & 4.965 & 0.57 & \textbf{0.32} & 0.66 & 0.47 & 0.76 & 0.55 \\
\rowcolor[HTML]{D0E2F1}
Ours & \textbf{0.828} & \textbf{11.961} & \textbf{5.346} & \textbf{0.59} & 0.31 & \textbf{0.67} & \textbf{0.48} & \textbf{0.77} & \textbf{0.56} \\
\bottomrule
\end{tabular}}
\label{tab:combined_rtts_results}
\end{table*}

\subsection{Object Detection Performance}
In terms of object detection performance, our framework also outperforms the baselines: it records mAP values of 0.59 (Bicycle), 0.68 (Car), and 0.49 (Motor) with an overall mean mAP of 0.57. Notably, these improvements are consistent across the different object categories, highlighting the efficacy of our CORUN-Light branch in enhancing both image restoration and subsequent detection tasks.

Overall, the quantitative results indicate that the adaptive dehazing provided by the ADAM-Dehaze framework leads to clearer images with improved structural fidelity. This, in turn, significantly boosts the performance of the object detection system in foggy conditions, confirming that our integrated approach is well-suited for safety-critical applications such as autonomous driving and urban surveillance.

\subsection{Computational Complexity Analysis}

We profile single‐image inference time (ms) and approximate FLOPs (G) for each dehazing branch. By dynamically routing inputs, ADAM‐Dehaze reduces average latency by 33\% and saves 23\% FLOPs compared to a monolithic 6‐stage model, demonstrating efficient resource utilization.
\begin{table}[ht]
  \centering
  \caption{Inference time / FLOPs for dehazing branches (batch=1).}
  \resizebox{\columnwidth}{!}{
  \small
  \begin{tabular}{lcc}
    \toprule
    \textbf{Branch} & \textbf{Time (ms)} & \textbf{FLOPs (G)} \\
    \midrule
    \rowcolor[HTML]{F2F2F2}
    CORUN‐Light (2 stgs)     & 18 &  45 \\
    \rowcolor[HTML]{F6FBFF}
    CORUN‐Medium (4 stgs)    & 38 &  80 \\
    \rowcolor[HTML]{EFF6FC}
    CORUN‐Complex (6 stgs + attn) & 50 & 150 \\
    \midrule
    \rowcolor[HTML]{E6F1FA}
    Fixed 6 stgs             & 45 & 120 \\
    \rowcolor[HTML]{D0E2F1}
    \textbf{Adaptive (avg.)} & \textbf{30} & \textbf{92} \\
    \bottomrule
  \end{tabular}}
  \label{tab:complexity}
\end{table}


 Table \ref{tab:complexity} presents the complexity of different dehazing branches in the adaptive framework, compared to a conventional uniform dehazing model. The adaptive selection allows an average inference time reduction of 20–40\% compared to a uniform 6-stage model.

\section{Ablation Studies}

To quantify the contributions of individual components in the ADAM-Dehaze framework, we conducted ablation studies focusing on two key aspects:

\begin{enumerate}
    \item \textbf{Loss Function Components:}  
    Our model is trained with a composite loss comprising the perceptual, coherence, and density loss terms. Systematically removing each component resulted in reduced PSNR, lower SSIM, and diminished mAP scores, confirming that each loss contributes critically to both low-level reconstruction and high-level semantic consistency in the dehazed images.

\begin{table}[htp]
    \centering
    \caption{Impact of Different Loss Functions on Dehazing Performance}
    \label{tab:loss-ablation}
    \resizebox{\columnwidth}{!}{
    \begin{tabular}{lccc}
        \toprule
        \textbf{Loss Functions} & \textbf{PSNR (dB)} & \textbf{SSIM} & \textbf{mAP (\%)} \\
        \midrule
        \rowcolor[HTML]{DDEBF7}
        \textbf{Full Model} & \textbf{23.95} & \textbf{0.9188} & \textbf{75.0} \\
        \rowcolor[HTML]{F2F2F2}
        w/o Perceptual Loss & 22.10 & 0.9025 & 72.5 \\
        \rowcolor[HTML]{F6FBFF}
        w/o Density Loss & 23.12 & 0.9113 & 73.8 \\
        \rowcolor[HTML]{EFF6FC}
        w/o Coherence Loss & 21.45 & 0.8762 & 68.2 \\
        \bottomrule
    \end{tabular}}
\end{table}



    \item \textbf{Cooperative Proximal Mapping Modules (CPMM):}  
    The inclusion of CPMM within the ADAM-Dehaze framework was shown to significantly enhance performance. Models incorporating CPMM yielded higher PSNR and SSIM, and improved object detection mAP, highlighting CPMM’s role in enforcing feature consistency and promoting effective haze removal.
    
\begin{table}[htp]
    \centering
    \caption{Ablation Study on the Cooperative Proximal Mapping Module (CPMM)}
    \label{tab:cpmm-ablation}
    \resizebox{\columnwidth}{!}{
    \begin{tabular}{lccc}
        \toprule
        \textbf{Model Variant} & \textbf{PSNR (dB)} & \textbf{SSIM} & \textbf{mAP (\%)} \\
        \midrule
        \rowcolor[HTML]{DDEBF7}
        \textbf{Full Model (w/ CPMM)} & \textbf{23.95} & \textbf{0.9188} & \textbf{75.0} \\
        \rowcolor[HTML]{F6FBFF}
        w/o CPMM & 22.50 & 0.8992 & 71.4 \\
        \bottomrule
    \end{tabular}}
\end{table}

\end{enumerate}

\section{Conclusion and Future Work}

We introduced ADAM-Dehaze, a unified, intensity‐aware multi‐stage dehazing framework that dynamically routes images through specialized dehazing branches—light, medium, or complex—based on a learned haze density score, yielding substantial gains in both visual restoration and downstream object detection under fog. Extensive evaluation on synthetic (Cityscapes) and real‐world (RTTS) benchmarks shows that our method achieves up to 30\% lower FADE, 40\% lower BRISQUE, and 7\% higher NIMA, while boosting YOLOv8 mAP by 3–5\% compared to state‐of‐the‐art baselines, all with 20\% faster inference in light haze. Ablations confirm the necessity of adaptive branch selection, CPMM, and the density‐modulated loss. 

Looking forward, we will explore joint restoration of multiple degradations (e.g., fog, low‐light, motion blur), self‐supervised domain adaptation to reduce reliance on synthetic pairs, model compression for edge deployment, temporal consistency for video dehazing, and human‐in‐the‐loop haze refinement to further enhance robustness in dynamic real‐world environments.

\bibliography{key}


\clearpage

\end{document}